\documentclass{article}

    \PassOptionsToPackage{round}{natbib}

    \usepackage[preprint]{neurips_2024}

\usepackage[utf8]{inputenc} 
\usepackage[T1]{fontenc}    
\usepackage{hyperref}       
\usepackage{url}            
\usepackage{booktabs}       
\usepackage{amsfonts}       
\usepackage{pifont}
\usepackage{bbold}
\usepackage{bbding}
\usepackage{multicol} 
\usepackage{multirow}
\usepackage{amssymb}
\usepackage{mathtools}
\usepackage{amsthm}
\usepackage{nicefrac}       
\usepackage{microtype}      
\usepackage{xcolor}         
\usepackage{cleveref}
\usepackage{algorithm}  
\usepackage{algorithmicx}
\usepackage{algpseudocode}  
\usepackage{graphicx}
\usepackage{subcaption}
\usepackage{caption}
\usepackage{soul}
\usepackage{array}
\usepackage{wrapfig}
\usepackage{amsmath}
\usepackage{mathtools}
\newtheorem{theorem}{Theorem}

\newtheorem{definition}{Definition}
\definecolor{magiccolor}{RGB}{205, 232, 248}
\newcommand{\cmark}{\ding{51}} 
\newcommand{\xmark}{\ding{55}} 

\hypersetup{
    hypertex=true,
    colorlinks=true,
    linkcolor=orange,
    anchorcolor=orange,
    citecolor=orange
}

\title{Q-WSL: Optimizing Goal-Conditioned RL with Weighted Supervised Learning via Dynamic Programming}

\author{Xing Lei \\
Xi’an Jiaotong University \\
\texttt{leixing@stu.xjtu.edu.cn} \\
\And
Xuetao Zhang \\
Xi’an Jiaotong University \\
\texttt{xuetaozh@xjtu.edu.cn} \\
\And
Zifeng Zhuang \\
Westlake University \\
\texttt{zhuangzifeng@westlake.edu.cn} \\
\And
Donglin Wang \\
Westlake University \\
\texttt{wangdonglin@westlake.edu.cn} \\
}

\begin{document}

\maketitle

\begin{abstract}
A novel class of advanced algorithms, termed Goal-Conditioned Weighted Supervised Learning (GCWSL), has recently emerged to tackle the challenges posed by sparse rewards in goal-conditioned reinforcement learning (RL). GCWSL consistently delivers strong performance across a diverse set of goal-reaching tasks due to its simplicity, effectiveness, and stability. However, GCWSL methods lack a crucial capability known as trajectory stitching, which is essential for learning optimal policies when faced with unseen skills during testing. This limitation becomes particularly pronounced when the replay buffer is predominantly filled with sub-optimal trajectories. In contrast, traditional TD-based RL methods, such as Q-learning, which utilize Dynamic Programming, do not face this issue but often experience instability due to the inherent difficulties in value function approximation.
In this paper, we propose Q-learning Weighted Supervised Learning (Q-WSL), a novel framework designed to overcome the limitations of GCWSL by incorporating the strengths of Dynamic Programming found in Q-learning. Q-WSL leverages Dynamic Programming results to output the optimal action of (state, goal) pairs across different trajectories within the replay buffer. This approach synergizes the strengths of both Q-learning and GCWSL, effectively mitigating their respective weaknesses and enhancing overall performance.
Empirical evaluations on challenging goal-reaching tasks demonstrate that Q-WSL surpasses other goal-conditioned approaches in terms of both performance and sample efficiency. Additionally, Q-WSL exhibits notable robustness in environments characterized by binary reward structures and environmental stochasticity. 
\end{abstract}

\section{Introduction}
\label{sec:introduction}

Deep reinforcement learning (RL) has demonstrated remarkable success in enabling agents to achieve complex objectives across a wide range of challenging and uncertain environments, including computer games \citep{hou2022parallel, liu2023exploring, kim2024dynasti}, robotic control \citep{xu2022learning, bouktif2023deep, xiao2024reinforcement}, and natural language processing \citep{sharifani2023machine, zeng2024rtrl}. Despite these advances, a fundamental challenge in deep RL lies in ensuring sample-efficient learning in environments with sparse rewards. This problem is particularly pronounced in goal-conditioned reinforcement learning (GCRL) \citep{kaelbling1993learning, schaul2015universal, andrychowicz2017hindsight, liu2022goal}, where agents must learn generalizable policies capable of reaching diverse goals.

Recently, we note that several goal-conditioned weighted supervised learning (GCWSL) methods have been proposed \citep{ghosh2021learning,yang2022rethinking,ma2022offline,hejna2023distance}  to tackle the GCRL challenges. In contrast to conventional RL approaches that maximizes discounted cumulative return, GCWSL provides
theoretical guarantees that supervised learning from hindsight relabeled experiences optimizes a lower bound on the goal-conditioned RL objective. Specifically, these methods first retrospectively infer which goals would have been fulfilled by a trajectory, irrespective of whether it successfully reached the desired goal. After this hindsight relabeling, GCWSL aims to imitate the corresponding sub-trajectories of (state, goal) pairs within the same trajectory in the relabeled data. Due to the benefit of the weighted function, GCWSL, unlike behavior cloning, can identify potentially optimal trajectories. Compared with other goal-conditioned RL or self-supervised (SL) methods
\citep{ding2019goal,chen2020learning,lynch2020learning,paster2020planning,eysenbach2020c,ghosh2021learning,eysenbach2022contrastive},
GCWSL has demonstrated outstanding performance across various goal-reaching tasks in a simple, effective, and stable manner.

Despite the successful application of GCWSL in effectively learning sparse rewards for certain goal-reaching tasks, 
some studies \citep{brandfonbrener2022does,yang2023swapped,ghugare2024closing} indicate that GCWSL may leads to sub-optimal policies when applied to the corresponding sub-trajectories of (state, goal) pairs across different trajectories and identify this issue as lacking of the ability to stitch trajectories. To effectively address this challenge, it is crucial to amalgamate information from multiple trajectories. On the other hand, we demonstrate that Q-learning possesses this ability due to its Dynamic Programming framework, leading to better policy outcomes.
However, Q-learning faces certain challenges in environments with sparse rewards. One of the key issues is the difficulty in propagating the value function back to the initial state, primarily due to the use of function approximation
\citep{Sutton1998}.

In this work, we introduce Q-WSL, a novel method designed to address the aforementioned challenges by leveraging Dynamic Programming within Q-learning to enhance the performance of GCWSL. Unlike many goal-conditioned RL algorithms that propose new architectures for agents to achieve better performance, Q-WSL focuses on improving the quality of the replay buffer, enabling more effective utilization of existing goal-conditioned RL algorithms.
Our approach is motivated by the following insight: while GCWSL often learns suboptimal policies for certain sub-trajectories of (state, goal) pairs across different trajectories, Q-learning can derive better actions. However, Q-learning tends to struggle when dealing with states that require backpropagation over large time steps. In these cases, GCWSL offers a distinct advantage by circumventing the need for backpropagation. By combining the strengths of both methods, Q-WSL mitigates their respective weaknesses, leading to a more robust and efficient performance.
To evaluate the effectiveness of Q-WSL, we conduct experiments using challenging GCRL benchmark environments. The results demonstrate that Q-WSL significantly improves sample efficiency compared to previous GCWSL approaches and other competitive GCRL methods, including DDPG+HER \citep{andrychowicz2017hindsight}, ActionModels \citep{2021Actionable}, and Model-based HER \citep{yang2021mher}. Additionally, Q-WSL exhibits strong robustness to variations in reward functions and environmental stochasticity.

We briefly summarize our contributions:
\begin{itemize}
    \item While state-of-the-art GCWSL methods tend to yield suboptimal performance for certain unseen skills during testing and lack trajectory stitching capabilities, we demonstrate that Q-learning inherently possesses this ability and can produce optimal actions in such scenarios.
    \item We introduce a novel and efficient GCRL method, Q-WSL, which harnesses the stitching capability of Q-learning to overcome the limitations of GCWSL. Additionally, we demonstrate that the optimization lower bound of Q-WSL exceeds that of several competitive GCRL algorithms, including WGCSL \citep{yang2022rethinking} and GCSL \citep{ghosh2021learning}.
    \item Experimental results on several GCRL benchmarks show that Q-WSL obtains better generalization performance and sample efficiency compared to prior GCRL methods. Additionally, Q-WSL is robust to variations in reward forms and environmental stochasticity.
\end{itemize}

\section{Related Work}

\paragraph{The Stitching Property} The concept of stitching, as discussed by \citep{ziebart2008maximum}, is a characteristic feature of TD-based RL algorithms, such as those described by \citep{mnih2013playing}, \citep{lillicrap2015continuous}, \citep{fujimoto2018addressing}, and \citep{kostrikov2021offline}, which employ Dynamic Programming techniques. This feature allows these algorithms to integrate data from different trajectories, thereby enhancing their effectiveness in managing complex tasks by leveraging historical data \citep{cheikhi2023statistical}. Recent works have shown that self-supervised learning also exhibits such property, particularly through methods like data augmentation
\citep{shorten2019survey,ghugare2024closing}
. However, there remains a significant performance gap between these methods and algorithms that incorporate Dynamic Programming.

This property motivates us to employ simple yet efficient Q-learning in TD-based RL to reach previously unvisited goals across different trajectories by leveraging Dynamic Programming. However, it is not clear that whether Q-learning process this important stitching property in GCRL. 
Using a straightforward counterexample, we demonstrate that stitching enables a unique property: the ability to infer solutions for composable goal-reaching trajectories during testing time. This includes maneuvering through specific (state, goal) pairs that never appear together during training, although they do appear separately.

\paragraph{Goal-conditioned Weighted Supervised Learning} Goal-conditioned weighted supervised learning methods (GCWSL)  \citep{liu2021self,yang2022rethinking,ma2022offline,hejna2023distance} are new outstanding goal-conditioned algorithms which enable agent to reach multiple goals in GCRL. These methods repeatedly perform imitation learning on self-collected, relabeled data without any RL updates. Furthermore, these methodologies undergo theoretical validation to confirm that self-supervised learning from relabeled experiences optimizes a lower bound on the GCRL objective. This validation guarantees that the learning process is both efficient and underpinned by a rigorous theoretical foundation.

However, recent researches \citep{brandfonbrener2022does,yang2023swapped,ghugare2024closing} indicate that these methods are sub-optimal for some unseen skills and lack the ability to stitch information from multiple trajectories. In response, our method, Q-WSL, which extends the framework of GCWSL by leveraging Dynamic Programming in Q-learning, can specifically designed to achieve optimal policy outcomes and integrates the capability for trajectory stitching.

\section{Preliminaries}
\label{sec:preliminaries}

\subsection{Goal-conditioned Reinforcement Learning} \label{sc:3.1}
Goal-conditioned Reinforcement Learning (GCRL) can be characterized by the tuple $\left\langle {\mathcal{S},\mathcal{A},\mathcal{G},\mathcal{P},r,\gamma,\rho_{0},T} \right\rangle$,
where 
$\mathcal{S}$,
$\mathcal{A}$,
$\mathcal{G}$,
$\gamma$,
$\rho_{0}$
and
$T$
respectively represent the state space,
action space,
goal space,
discounted factor,
the distribution of initial states and the horizon of the episode.
$\mathcal{P}:\mathcal{P}(s^{\prime}|s,a)$
is the dynamic transition function,
and $r:r(s,a,g)$ is typically a simple unshaped binary signal.
The sparse reward function employed in GCRL can be expressed as follows:
\begin{align} \label{eq:1}
r(s_t,a_t,g)=\begin{cases}0,&\|\phi(s_t)-g\|_2^2<\delta\\-1,&\text{otherwise}\end{cases}, &
\end{align}
$\delta$ is a threshold and
$\phi:\mathcal{S}\rightarrow \mathcal{G}$
is a known state-to-goal mapping function from states to goals.
Agents must develop a policy $\pi:\mathcal{S}\times \mathcal{G}\rightarrow \mathcal{A}$ that optimizes the expected returns associated with reaching goals sampled from the goal distribution $p(g)$:
\begin{equation} \label{eq:2}
\mathcal{J}(\pi)=E_{g\sim p(g),a_t\sim\pi,s_{t+1}\sim p(\cdot|s_t,a_t)}\left[\sum_{t=0}^{T}\gamma^tr(s_t,a_t,g)\right].
\end{equation}
\paragraph{Q-learning} methods optimizes $\mathcal{J}(\pi)$ via the gradient computation:
\begin{equation} \label{eq:3}
\mathcal{J}(\pi)=\mathbb{E}_{g,a_t,s_{t+1}}\left[\sum_{t=0}^{T} Q^{\pi}(s_t,a_t,g)\right],
\end{equation}
where $Q^{\pi}(s_t,a_t,g)$ is the goal-conditioned Q-function.
Note that in $\mathbb{1}\left[\phi(s_t)=g\right]$ reward 
(i.e, If the state 
$s$ reaches the goal $g$, then the reward is 1; otherwise, it is 0.),
the $Q^{\pi}(s_t,a_t,g)$ is always zero.
This Q-function is not useful for predicting the future goal distribution.

\subsection{Goal-conditioned Weighted Supervised Learning} \label{sc:3.2}
In contrast to goal-conditioned RL methods, which focus on directly optimizing the discounted cumulative return,
Goal-Conditioned Weighted Supervised Learning methods (GCWSL) introduce an innovative learning framework by iteratively relabeling and imitating self-generated experiences. This iterative process enhances the model's ability to generalize and efficiently learn from past experiences, thereby improving its overall performance in goal-reaching tasks.
Trajectories in the replay buffer are relabeled with
hindsight method
\citep{andrychowicz2017hindsight}.
And then the policy optimization satisfies the following definition during imitating:
\begin{definition}
Given replay buffer $\mathcal{B}$ and hindsight relabeling mechanism,
the objective of GCWSL is to mimic those relabeled transitions through supervised learning with weighted function: 
\begin{equation} \label{eq:4}
\mathcal{J}_{GCWSL}(\pi) = \mathbb{E}_{(s_{t},a_{t},g)\sim \mathcal{B}_{r}}\left[f(A)\cdot\log\pi_{\theta}(a_{t}|s_{t},g)\right],
\end{equation}     
\end{definition}
where $\mathcal{B}_{r}$ denotes relabeled replay buffer,
$g=\phi(s_i)$ denotes the relabeled goals
for
$i\geq t$
\begin{wrapfigure}[9]{R}{0.6\textwidth}
    \vspace{-1.5em}
    \centering
    \begin{minipage}{0.6\textwidth}
        \begin{algorithm}[H]
           \caption{Goal-conditioned self-supervised learning with weight function} 
           \label{alg:WBC}
           \begin{algorithmic}[1]
              \State \textbf{Input} behavior policy $\beta(a\mid s,g)$
              \State $\mathcal{B}_r \leftarrow$ RELABEL ($\tau$) where $\tau \sim \beta(\tau)$
              \State $\mathcal{J}_{GCWSL}(\pi)\leftarrow\mathbb{E}_{(s_t,a_t,g)\sim \mathcal{B}_r}\left[f(A) \cdot\log\pi(a_t\mid s_t,g)\right]$
              \State $\pi(a\mid s,g)\leftarrow\arg\max_{\pi}\mathcal{J}_{GCWSL}(\pi)$
              \State \textbf{return} $\pi(a\mid s,g)$
           \end{algorithmic} 
        \end{algorithm}
    \end{minipage}
\end{wrapfigure}
and $f(A)$ is a function about advantage $A:=A(s_t,a_t,g)$.
The weighted function $f(A)$ exists various forms in GCWSL methods
\citep{ghosh2021learning,yang2022rethinking,ma2022offline,hejna2023distance}.
(See \Cref{sc:a.2} for the specific weight function meaning).
Therefore GCWSL includes typical two process,
acquiring sub-trajectories corresponding to state-goal
(i.e, $(s,g)$) pairs and imitating them.
In the process of imitation,
GCWSL first train the related specific weighted function,
and then extract the policy with the \Cref{eq:4}.

We briefly summarize the GCWSL training procedure in
\Cref{alg:WBC},
where $\beta(a|s,g):=\pi_{old}(a|s,g)$ in practice. And then we have following theorem.
\begin{theorem} \label{theorem:3.2}
Suppose there exist a deterministic policy $\pi_{relabel}$ capable of generating relabeled data $\mathcal{B}_r$,
GCWSL is a variant form of goal-conditioned AWR over $\mathcal{B}_{r}$:
\begin{equation} \label{eq:5}
\pi_{k+1}=\underset{\pi\in\Pi}{\operatorname*{\arg\max}}\mathbb{E}_{(s,g,s')\sim \mathcal{B}_r,a_t\sim\pi(s,g)}\left[A^{\pi_k}(s,a,g)\right],\text{s.t.}~\mathcal{D}_{\mathrm{KL}} \left(\pi\|\pi_{relabel}\right)<\epsilon,
\end{equation}
\end{theorem}
where $A^{\pi_k}=Q^{\pi_k}(s,a,g) - V^{\pi_k}(s,g)$,
$\mathcal{D}_{\mathrm{KL}}$ is the KL-divergence and $g$ is the goals from relabeled data. See \Cref{sc:a.1} for detail proof. This theorem indicate GCWSL guarantees policy improvement when learning from the relabelled data.

\section{Q-WSL: Q-learning Weighted Supervised Learning}
To generalize GCWSL’s skills outside the relabeled data and further improve sample efficiency,
we start with a counter example to illustrate why Q-learning in TD-based RL can solve the experience stitching problem in \Cref{sec:4.1}.
Then we propose a modified of GCWSL called Q-WSL that leveraging the stitching ability of Q-learning in \Cref{sec:5.1}. 
In \Cref{sec:5.1},
we also provide theoretical guarantee for Q-WSL.
Further,
we introduce the overall algorithm implementation process of Q-WSL in \Cref{sec:5.2}.
Finally, we summarize and compare Q-WSL with previous related work in \Cref{sec:5.3}.
\begin{figure*}[h!]
   \centering
   \includegraphics[width=\textwidth]{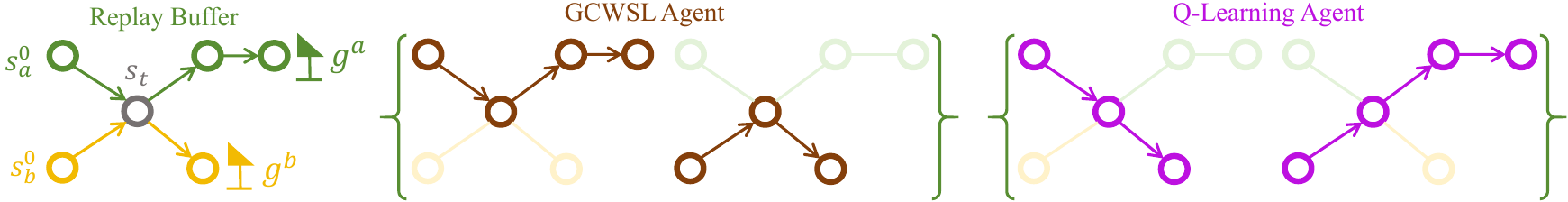}
   \caption{A counterexample of GCWSL lack stitching ability while Q-learning can output the path across different trajectories.
   $g\in \left[g^{a},g^{b} \right]$ are conveniently expressed as reachable goals.
   Specifically,
   the off-policy replay buffer stores history data collected by behavior policy $\beta(a|s,g)$ during training,
   and then $f(A)\cdot\log\pi(a|s,g)$ is maximized with batch
   (state, goal) pairs from this buffer.
   During testing,
   the $\pi(a|s,g)$ is tested on random batch (state, goal) pairs.
   GCWSL can consistently learn (state, goal) pairs within the same trajectory (i.e, $(s_0^a,g^a),(s_0^b,g^b)$), which are represented by the same color. However, it lacks the guarantee of generating the correct actions for (state, goal) pairs that originate from different trajectories (i.e, $(s_0^a,g^b),(s_0^b,g^a)$).
   In contrast,
   Q-learning can output optimal actions for these (state, goal) pairs across different trajectories thanks to Dynamic Programming.
   }
   \label{fig:weighted_trajectory}
\end{figure*}

\subsection{Trajectory Stitching Example} \label{sec:4.1}
From the perspective of GCWSL updating the corresponding 
$(s,g)$ pairs, GCWSL aims to identify and imitate the optimal sub-trajectory from the combination data generated by the old policy that meets the 
$(s,g)$ pairs. The optimal trajectory here satisfies the weighted function $f(A)$, such as the 
the horizon length from $s$ to $g$ 
in \citep{yang2022rethinking}.
However, it has been observed that when data is aggregated from a variety of older policies, certain 
$(s,g)$ pairs are rarely encountered within the same trajectory, despite being common in different trajectories. In these instances, GCWSL often struggles to produce an optimal policy.

Consider an example illustrated in \Cref{fig:weighted_trajectory}, where the goals 
$g\in \left[g^{a},g^{b} \right]$ are reachable from the states 
$s\in \left[s_0^a,s_0^b,s_t \right]$ although not explicitly present in the off-policy replay buffer. During training, GCWSL is updated with (state, goal) pairs sampled from the same trajectory, such as 
$(s_0^a,g^{a})$ and $(s_0^b,g^{b})$. However, during testing, when unseen (state, goal) pairs such as $(s_0^a,g^{b}),(s_0^b,g^{a})$ originating from distinct trajectories are introduced, GCWSL cannot find the correct actions
(proof can be found in Figure 1 and Lemma 4.1 of
\citep{ghugare2024closing}).
In contrast, TD-based RL methods such as Q-learning can output the correct actions for $(s_0^a,g^{b}),(s_0^b,g^{a})$ due to the Dynamic Programming component. For instance, consider a value estimation example, $V (s_t, g) = r + V (s_{t+1}, g)$.
For the pair $(s_{t+1}, g)$, provided that $g$ is accessible and the value $V (s_{t+1}, g)$ is accurately approximated, Q-learning can back-propagate the state-goal (state-action-goal) values to preceding pairs. An example from the off-policy replay buffer illustrates the state $s_t$ as a "hub state" shared by two initial trajectories, from which all goals $g^i\sim[g^a,g^b]$ are reachable. The values of these goal-conditioned states $Q(s_t,a_t^i,g^i)$ can be computed and recursively back-propagated to $Q(s_{0},a_{0}^{i},g^{i}),i\in \left[a,b \right]$. Q-learning has the capability to propagate rewards through paths that are connected retrospectively. This allows Q-learning to determine the optimal actions for state-goal pairs that have not been encountered together previously, such as $(s_0^a,g^{b})$ and $(s_0^b,g^{a})$. By iteratively back-propagating value estimates through all potential trajectory combinations, Q-learning ensures that values are consistently updated, thereby facilitating the correct action selection.
In summary, Q-learning can link (state, goal) pairs across different trajectories, a capability that GCWSL lacks.

\subsection{Bounding KL-Constrained Values and Q-WSL} \label{sec:5.1}
If we he replace the advantage function $A^{\pi}(s,a,g)$ in 
\Cref{eq:5}
by Q-function $Q^{\pi}(s,a,g)$
we have the policy improvement objective:
\begin{equation} \label{eq:16}
\pi_{k+1}=\arg\max_{\pi}\mathbb{E}\left[Q^{\pi_{k}}(s,\pi_{k}(s,g),g)\right],\nonumber\\
\;\int_{a}\pi(a|s,g)\boldsymbol{d}a=1,\;\mathbf{s.t.}~\operatorname{KL}(\pi\|\pi_{relabel})\leq\epsilon.
\end{equation}
Since minimizing the KL-divergence is equivalent to maximizing the likelihood \citep{lecun2015deep},
we have the following Lagrangian:
\begin{equation} \label{eq:17}
\mathcal{L}(\lambda,\pi^H)=\mathbb{E}_{a\sim\pi(\cdot|s,g)}\left[Q^{\pi}(s,a,g)\right]+ \nonumber \\
\lambda\mathbb{E}_{a,s,g\sim \mathcal{B}_{r}}\log(\pi(a|s,g)).
\end{equation}
For a deterministic policy 
$\pi$, it can be regarded as a Dirac-Delta function. Consequently, the constraint $\int_{a}\pi(a|s,g)\boldsymbol{d}a\\=1$ is always satisfied. Therefore, \Cref{eq:17} simplifies to:
\begin{equation} \label{eq:18}
\underset{\pi}{\arg\min}\mathbb{E}_{a,s,g\sim \mathcal{B}_{r}}\left[-Q^{\pi}(s,a,g)+\|\pi(s,g)-a\|_2^2\right].
\end{equation}
It also can be rewritten to maximize the objective form:
\begin{equation}
\mathcal{J}_{q}(\pi)=\mathbb{E}_{s,a,g\sim {\mathcal{B}_r}}\left[Q^{\pi}(s,a,g)+\log(\pi(a|s,g))\right].   
\end{equation}
The two objectives optimize the policy through different mechanisms. The objective 
$\mathcal{J}_{wbc}(\pi)$ represents weighted behavior cloning, which is typically used in prior GCWSL methods. However, 
$\mathcal{J}_{wbc}(\pi)$ may not yield the optimal goal-conditioned actions ( \Cref{sc:3.2}), since the policy constraint is inclined to be more conservative, it tends to align with the actions present in the dataset. On the other hand, the objective $\mathcal{J}(\theta)_{q}$ optimizes the policy through Q-learning, a method inherent to TD-based RL, as corroborated by \citep{fujimoto2021minimalist}.

It is important to note that these two objectives are mutually reinforcing: the weighted behavior cloning guarantees that the policy constraint remains adaptable and selects high-quality samples for the $(s,g)$ pair corresponding to sub-trajectories. Concurrently, the Q-learning steers the policy towards the optimal action for the $(s,g)$ pair.

Based on this discussion and the interoperation of these two objectives, we combine \Cref{eq:16} with \Cref{eq:5} to optimize the policy concurrently. This leads to our Q-WSL policy optimization objective for policy updates:
\begin{equation} \label{eq:20}
\mathcal{J}_{Q-WSL}=\mathbb{E}_{(s,a,g)\sim \mathcal{B}_r}\big[\underbrace{Q^{\pi}(s,a,g}_\text{Q-learning})
+\eta\underbrace{f(A^{\pi}(s,a,g))\cdot\log\pi_{\theta}(s,a,g)}_\text{Advantage-weighted Regression}\big],
\end{equation}
where $\eta$ is defined to balance the RL term
( maximizing Q-function)
and imitation term
( maximizing the weighted behavior cloning term)
like TD3+BC \citep{fujimoto2021minimalist}.

\subsection{Practice Algorithm} \label{sec:5.2}
We adopt the DDPG+HER as our baseline framework following
\citep{fang2019curriculum,yang2021mher},
therefore for a deterministic policy $\pi(s,g)$,
the \Cref{eq:13} and \Cref{eq:18} are used during training procedure. 
Moreover,
we follow WGCSL 
\citep{yang2022rethinking} to define $f(A)$ equal to $\gamma^{i-t}\exp_{clip}(A(s_t,a_t,g))\cdot\epsilon(A(s_{t},a_{t},g))$,
where $A(s_t,a_t,g) = r(s_t,a_t,g)+\gamma V(s_{t+1},g)-V(s_t,g)$ and
the value function $V(s_t,g)$ is estimated using Q-value function as $V(s_t,g)=Q(s_t,\pi(s_{t+1},g),g)$ 
Because in the experiment, 
we found that the effect of this form of $f(A)$ is better for online goal-conditioned RL, 
and other forms of $f(A)$ may be more friendly for offline settings.
The final overview of Q-WSL is presented in \Cref{alg:Q-WSL}.\\
\paragraph{Compared with WGCSL} Motivated by \citep{2021Goal,yang2022rethinking}, 
we also compared with WGCSL and have the following boundary:
\begin{theorem} \label{theorem:5.1}
Suppose function
$\exp_{clip}(
A(s_{t},a_{t},g)\cdot\epsilon(A(s_{t}$
$,a_{t},g))\geq1, g = \phi(s_i), i\geq1$ over the state-action-goal combination.
Consider a finite-horizon discrete Markov Decision Process (MDP), a deterministic policy $\pi$ that selects actions with non-zero probability, and a sparse reward function 
$r(s_t, a_t, g) = \mathbb{1}\left[\phi(s_i) = g\right]$, where 
$\mathbb{1}\left[\phi(s_i) = g\right]$ is an indicator function. Given trajectories $\tau = (s_1,a_1,\cdots,s_T,a_T)$ and a discount factor 
$\gamma \in (0, 1]$, the following bounds hold:
\begin{flalign}
&\
\mathcal{J}_{Q-WSL}(\pi) \geq \mathcal{J}_{WGCSL}(\pi)\geq \mathcal{J}_{GCSL}(\pi).
&
\end{flalign}
\end{theorem}
We defer the proof to \Cref{sc:a.1}.
This theorem indicates that under a reward of $r(s_t, a_t, g) = \mathbb{1}\left[\phi(s_i) = g\right]$, 
the optimization objective of Q-WSL is at least better than that of WGCSL and GCSL.
\begin{algorithm*}[h]   
   \caption{Q-WSL for Goal-conditioned RL} 
   \label{alg:Q-WSL}
   \begin{algorithmic}[1]  
      \State \textbf{Initialize} off-policy replay buffer $\mathcal{B}$
      \State \textbf{Initialize} policy $\pi_{\theta}$,
      policy target network weights $\bar{\theta}\leftarrow\theta$,
      $Q$-network $Q_\psi$,
      $Q$-target network weights $\bar{\psi}\leftarrow\psi$,
      \While{a fixed number of iteration}
        \State Collect trajectories with the policy $\pi_{\theta}$ and save to the replay buffer $\mathcal{B}$
        \For{Update goal-conditioned policy step}
            \State Generate relabeled data $\mathcal{B}_r$ from $\mathcal{B}$:
            $m\leftarrow \{(s_t,a_t,g,r(s_t,a_t,g)),i\geq t\}$,
            where $g=\phi(s_{i}), i\geq t$
            \State \textcolor{orange}{// Sample relabeled data and calculate Q-value function}
            \State Calculate $Q$ targets $y_q$:
            $y_{q}=r(s_{t},a_{t},g)+\gamma Q_{\bar{\psi}}(s_{t+1},\pi_{\theta}(s_{t+1},g),g)$
            \State Clip the $Q$ function:
            $y_{q}=y_{q}.clip(-1/(1-\gamma),0)$ 
            \State Optimize $Q$ by minimizing following TD error:
            $\mathbb{E}_{(s_t,a_t,g)\sim B_r}\left[(y_q-Q_{\psi}(s_t,a_t,g))^2\right]$
            \State Calculate the weight within WGCSL:
            $w_A=\gamma^{i-t}\exp_{clip}(A(s_t,a_t,g))\cdot\epsilon(A(s_{t},a_{t},g))$
            \State \textcolor{orange}{//Policy optimization with Q-WSL}
            \State Update $\pi_{\theta}$ with maximize goal-conditioned policy  optimization with \Cref{eq:13} and \Cref{eq:18}
        \EndFor
        \State Soft update the policy target network and
               $Q$-target network:
        $\bar{\theta}\leftarrow\tau\theta+(1-\tau)\bar{\theta}$,
        $\bar{\psi}\leftarrow\tau\psi+(1-\tau)\bar{\psi}$
      \EndWhile
   \end{algorithmic} 
\end{algorithm*}

\subsection{Compared with Previous Goal-conditioned Methods} \label{sec:5.3}
\autoref{tb:comparsion} compares the policy optimization objectives of the proposed goal-conditioned methods with those of prior works. For this comparison, we selected four actor-critic methods \citep{silver2014deterministic,2021Actionable,andrychowicz2017hindsight,yang2021mher}, and four state-of-the-art goal-conditioned weighted supervised learning methods \citep{liu2021self,yang2022rethinking,ma2022offline,hejna2023distance}.

\autoref{tb:comparsion} demonstrates that most prior weighted self-supervised methods are based on advantage-weighted regression approaches, with the exception of GCSL \citep{liu2021self}, which employs an objective similar to \Cref{eq:4}. However, methodologies such as GCSL \citep{liu2021self}, WGCSL \citep{yang2022rethinking}, and DWSL \citep{hejna2023distance} are constrained to utilizing trajectories from successful examples only. GoFAR \citep{ma2022offline} requires the collected dataset to cover all reachable goals, and Action Models \citep{2021Actionable} fails on noisy trajectories.
In contrast, our Q-WSL avoids these drawbacks. Our approach shares the principles of prior actor-critic methods such as DDPG \citep{silver2014deterministic}, DDPG+HER \citep{andrychowicz2017hindsight}, and Model-based HER \citep{yang2021mher}. To properly recognize the contributions of previous studies, we assert that our proposed method capitalizes on the benefits of both Q-learning optimization and advantage-weighted regression. Consequently, many of the aforementioned techniques, particularly WGCSL, GoFAR, and DWSL, can be seamlessly integrated into our model.
\begin{table}[h]
    \renewcommand{\arraystretch}{2}
    \centering
    \caption{Comparison of various goal-conditioned methods policy objectives.
             $F$ signifies the advantage-weighted function in WGCSL,
             $F_{\star}'$ signifies the $f$-divergence function in GoFAR
             and $\{\mathcal{B},\mathcal{B}_r,\mathcal{B}_m\}$ correspond to 
             $\{$replay buffer, relabeled data, model-based relabeled data$\}$ respectively.}
    \label{tb:comparsion}
    \resizebox{\textwidth}{!}{
    \Huge
    \begin{tabular}{llcccc}
    \hline
    \multirow{2}{*}{\textbf{Goal-conditioned Methods}} & \multicolumn{3}{c}{\textbf{Optimization}}                          & \multirow{2}{*}{\textbf{Relabeling}} \\ \cline{2-4}
                              & \textbf{Policy objective} & \textbf{Q-learning} & \textbf{Advantage-weighted Regression} &                           &                          \\ \hline
    DDPG\citep{2018CONTINUOUS}                      & $\min_{\pi_{\theta}}\mathbb{E}_{(s,g)\sim \mathcal{B}}\left[-Q(s,\pi_{\theta}(s,g),g)\right]$                  & \scalebox{1.5}{\cmark}                 & \scalebox{1.5}{\xmark}                                                &\scalebox{1.5}{\xmark}                          \\
    Action Models\citep{chebotar2021actionable}                        & $\min_{\pi_{\theta}}\mathbb{E}_{(s,g,Q')\sim \mathcal{B}}\left[-Q'(s,\pi_{\theta}(s,g),g)\right]$                  & \scalebox{1.5}{\xmark}                 & \scalebox{1.5}{\xmark}                                                & \scalebox{1.5}{\xmark}                          \\
    DDPG+HER\citep{andrychowicz2017hindsight}                  & $\min_{\pi_{\theta}}\mathbb{E}_{(s,g)\sim \mathcal{B}_r}\left[-Q(s,\pi_{\theta}(s,g),g)\right]$                  & \scalebox{1.5}{\cmark}                 & \scalebox{1.5}{\xmark}                                                & \scalebox{1.5}{\cmark}                          \\
    Model-based HER\citep{yang2021mher}           & $\min_{\pi_{\theta}}\mathbb{E}_{(s,a,g)\sim \mathcal{B}_m}\left[-Q(s,\pi(s,g),g)\right]+\alpha\mathbb{E}_{(s,a,g)\sim \mathcal{B}_m}\left[\|a-\pi_{\theta}(s,g)\|_2^2\right]$                 & \scalebox{1.5}{\cmark}                 & \scalebox{1.5}{\xmark}                                                & \scalebox{1.5}{\cmark}                          \\
    GCSL\citep{ghosh2021learning}                      & $\min_{\pi_{\theta}}\mathbb{E}_{(s,a,g)\sim \mathcal{B}_r}\left[-\log\pi_{\theta}(a|s,g)\right]$                  & \scalebox{1.5}{\xmark}                 & \scalebox{1.5}{\xmark}                                                &\scalebox{1.5}{\cmark}                          \\
    WGCSL\citep{yang2022rethinking}                     & $\min_{\pi_{\theta}}\mathbb{E}_{(s_{t},a_{t},g)\sim \mathcal{B}_r}\left[-F(A(s_{t},a_{t},g),t)\cdot\log\pi_{\theta}(a_{t}|s_{t},g)\right]$                 & \scalebox{1.5}{\xmark}                 & \scalebox{1.5}{\cmark}                                               & \scalebox{1.5}{\cmark}                          \\
    GoFar\citep{ma2022offline}                     & $\min_{\pi_{\theta}}\mathbb{E}_{(s,a,g)\sim \mathcal{B}}\left[F_{\star}'(R(s;g)+\gamma\mathcal{T}\mathcal{V}^{*}(s,a;g)-\mathcal{V}(s;g))\log\pi_{\theta}(a|s,g)\right]$                 & \scalebox{1.5}{\xmark}                 & \scalebox{1.5}{\cmark}                                                & \scalebox{1.5}{\cmark}                          \\
    DWSL\citep{hejna2023distance}                      & $\min_{\pi_{\theta}}\mathbb{E}_{(s_t,a_t,g)\sim \mathcal{B}_r}\left[-e^{A/\omega}\cdot\log\pi_{\theta}(a_t|s_t,g)\right]$                  & \scalebox{1.5}{\xmark}                 & \scalebox{1.5}{\cmark}                                                & \scalebox{1.5}{\cmark}                          \\ \hline
    \textbf{Q-WSL(ours)}                & $\operatorname*{min}_{\pi_{\theta}}\mathbb{E}_{(s_t,a_t,g)\sim \mathcal{B}_r}\left[-Q(s_t,\pi_{\theta}(s_{t+1},g),g)+\eta F(A(s_t,a_t,g),t)\|\pi_{\theta}(s_t,g)-a\|_2^2\right]$                 & \scalebox{1.5}{\cmark}                 & \scalebox{1.5}{\cmark}                                               & \scalebox{1.5}{\cmark}                          \\ \hline
    \end{tabular}
    }
\end{table}

\section{Experiments}
\label{sec:experiments}

\begin{figure}[h]
    \centering
    \begin{minipage}{0.18\linewidth}
        \centering
        \includegraphics[height=2.5cm, width=2.5cm]{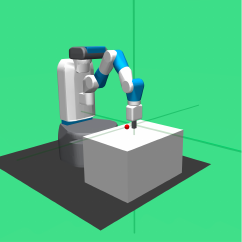}
    \end{minipage}
    \begin{minipage}{0.18\linewidth}
        \centering
        \includegraphics[height=2.5cm, width=2.5cm]{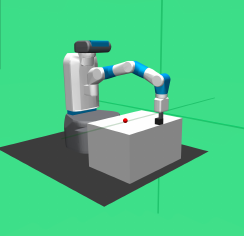}
    \end{minipage}
    \begin{minipage}{0.18\linewidth}
        \centering
        \includegraphics[height=2.5cm, width=2.5cm]{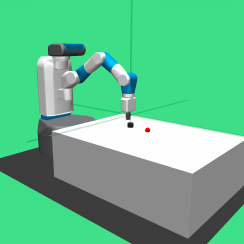}
    \end{minipage}
    \begin{minipage}{0.18\linewidth}
        \centering
        \includegraphics[height=2.5cm, width=2.5cm]{experiment_pdf/fetchslide.pdf}
    \end{minipage}
    \begin{minipage}{0.18\linewidth}
        \centering
        \includegraphics[height=2.5cm, width=2.5cm]{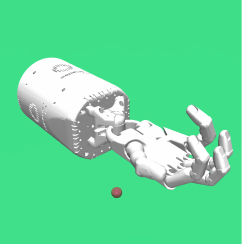}
    \end{minipage}
    \caption{Examples of goal-conditioned benchmark environments: (a) FetchReach, (b) FetchPush, (c) FetchSlide, (d) FetchPickAndPlace, (e) HandReach.}
    \label{fig:multi_goal tasks}
\end{figure}

We begin by introducing the benchmarks and baseline methods, followed by a detailed description of the experimental setup. Subsequently, we present the results and analysis, which confirm our assumptions and theoretical framework.\\
\paragraph{Benchmarks} We use some standard goal-conditioned RL research benchmarks \citep{plappert2018multi} including one task on the $Shadow-hand$ and four tasks on the $Fetch$ robot
(See \Cref{fig:multi_goal tasks}).
\paragraph{Baselines} 
In this section, we compare our method Q-WSL with prior works on various goal-conditioned policy learning optimization algorithms as show in
\autoref{tb:comparsion}
. We conducted a series of comparative experiments by implementing the baseline methods within the same framework as our proposed approach.
Detailed baselines algorithm implementation is shown in \Cref{sc:a.2}.
We provide a brief overview of our baseline algorithms as bellow:
\begin{itemize}
    \item \textbf{DDPG} \citep{2018CONTINUOUS}
    is an efficient off-policy actor-critic method for learning continuous actions. In this paper, we adopt it as the basic framework.
    \item \textbf{DDPG+HER} \citep{andrychowicz2017hindsight}
    combines DDPG with HER, which learns from failed experiences with sparse rewards.
    \item \textbf{Actionable Models} \citep{chebotar2021actionable}
    subtly combines goal-chaining with conservative $Q$-Learning.
    we re-implement this goal relabeling technique which apply the the Q-value at
    the final state of trajectories in $\mathcal{B}$ to enable goal chaining, as well as the negative action sampling operation.
    \item \textbf{MHER} \citep{yang2021mher}
    constructs a dynamics model using historical trajectories
    and combines current policy to generate future trajectories for goal relabeling.
    \item \textbf{GCSL} \citep{ghosh2021learning}
    is a typical goal-conditioned supervised learning.
    GCSL incorporates hindsight relabeling in conjunction with behavior cloning to imitate the suboptimal trajectory.
    \item \textbf{WGCSL} \citep{yang2022rethinking} introduces an enhanced version of GCSL, incorporating both discounted goal relabeling and advantage-weighted updates into the policy learning process. This integration aims to improve the efficiency and effectiveness of the policy learning updates.
    \item \textbf{GoFar} \citep{ma2022offline} employs advantage-weighted regression with $f$-divergence regularization based on state-occupancy matching. Additionally, it introduces a hierarchical framework in offline goal-conditioned reinforcement learning to effectively manage long-horizon tasks.
    \item \textbf{DWSL} \citep{hejna2023distance} initially creates a model to quantify the distance between state and goal. And then DWSL utilizes this learned distance to improve and extract policy.
\end{itemize}

\paragraph{Architecture Setup} We utilize the DDPG \citep{2018CONTINUOUS} in conjunction with HER as the foundational algorithm for goal-conditioned RL, against which we benchmark various baselines. Specifically, the critic architecture $Q(s, a, g)$ adheres to the Universal Value Function Approximators (UVFA) model \citep{schaul2015universal}. The actor network generates actions parameterized by a Gaussian distribution. The actor and critic networks are structured with three hidden layers of multi-layer perceptrons (MLPs), each comprising 256 neurons and employing ReLU activations. The architectural setup can be denoted as [linear-ReLU]×3, followed by a linear output layer. We follow the optimal parameter settings for DDPG+HER as specified in \citep{plappert2018multi}. All baseline hyperparameters are maintained consistently. Additional details are provided in \Cref{appendix:a.7}.

\paragraph{Evaluation Setup} For each baseline and task, we conducted evaluations using random five seeds
(e.g, $\{ 100, 200, 300, 400,500 \}$). The policy was trained for 1000 episodes per epoch. Upon completing each training epoch, the policy's performance was measured by calculating the average success rate from 100 independent rollouts, each using randomly selected goals. These success rates were averaged across five seeds and plotted over the learning epochs, with the standard deviation illustrated as a shaded region on the graph.
\subsection{Performance Evaluation on Multi-goal Benchmark Results} \label{sec:6.1}
For all baselines, we use 16 CPUs to train the agent for 50 epochs in all tasks. Upon completing the training phase, the most effective policy is evaluated by testing it on the designated tasks. The performance outcomes are then expressed as average success rates. Performance comparisons across training epochs are illustrated in \Cref{success table}. For Q-WSL, the policy objective parameter $\eta$ is set to 0.1.
\begin{table}[h]
\Large
\renewcommand{\arraystretch}{1.5}
\centering
\caption{Mean success rate (\%) for challenging multi-goal robotics tasks after training}
\label{success table}
\resizebox{\linewidth}{!}{
\begin{tabular}{l|l|llll|llll}
\hline
\textbf{Task}                       & \textbf{}     & \multicolumn{4}{l|}{\textbf{Weighted Behavior Cloning}}                      & \multicolumn{4}{l}{\textbf{Actor-Critic Methods}}                             \\
                                           & \textbf{Q-WSL(Ours)} & DWSL &GoFar & WGCSL & GCSL & DDPG+HER & MHER & Action Models & DDPG \\ \hline
FetchReach         & \colorbox{magiccolor}{100.0} $\pm$ 0.0              &96.0 $\pm$ 2.74                &100.0 $\pm$ 0.0                &99.0 $\pm$ 2.2              &93.2 $\pm$ 4.2              &99.2 $\pm$ 1.3              &99.8 $\pm$ 0.4                         &97.4 $\pm$ 5.2                      &100.0 $\pm$ 0.0      \\
FetchPush         &\colorbox{magiccolor}{99.4} $\pm$ 0.89               &7.2 $\pm$ 3.9                & 88.8 $\pm$ 8.0               &94.0 $\pm$ 5.3              &5.4 $\pm$ 4.8                &91.6 $\pm$ 17.3            
& 96.8 $\pm$ 4.5                         & 6.8 $\pm$ 3.5                      &7.0 $\pm$ 4.2     \\
FetchSlide         &\colorbox{magiccolor}{50.6} $\pm$ 8.4               &0.2 $\pm$ 0.45                &36.6 $\pm$ 10.9                &29.0 $\pm$ 14.6               &0.4 $\pm$ 0.5              &30.2 $\pm$ 14.9              &43.6 $\pm$ 8.56                          & 0.2 $\pm$ 0.45                      &2.2 $\pm$ 3.8      \\
FetchPickAndPlace          &\colorbox{magiccolor}{96} $\pm$ 2.9               & 3.8 $\pm$ 2.28               &45.2 $\pm$ 4.14                &66.4 $\pm$ 12.44              &3.8 $\pm$ 2.28              &92.6 $\pm$ 1.14       
&95.4 $\pm$ 2.88                         &3.4 $\pm$ 2.51                       &3.8 $\pm$ 2.3     \\
HandReach                         & \colorbox{magiccolor}{59.8} $\pm$ 11.8              &0.0 $\pm$ 0.0                &17.6 $\pm$ 4.56                &19.2 $\pm$ 10.59              &0.0 $\pm$ 0.0               &\colorbox{magiccolor}{64.8} $\pm$ 9.9              &0.0 $\pm$ 0.0                          &0.0 $\pm$ 0.0                                &0.0 $\pm$ 0.0      \\ \hline
\textbf{Average}                      & \colorbox{magiccolor}{81.16}               &21.44                &57.64                &61.52               & 20.56              & 75.68             & 67.12                         & 21.56                               & 22.6      \\ \hline
\end{tabular}
}
\end{table}
From \Cref{success table}, it is evident that Q-WSL attains significantly higher performance than prior baselines. The results indicate that DDPG and Actionable Models perform poorly across all tasks, while DDPG+HER, MHER, and GCWSL methods, which benefit from HER, demonstrate greater effectiveness. This underscores the importance of HER in multi-goal RL for enhancing learning with sparse rewards and improving sample efficiency. Furthermore, TD-based RL methods built upon HER consistently outperform GCWSL methods under the same conditions, confirming our conclusions in \Cref{sec:4.1}.

\Cref{success table} also shows that compared to the best baseline method, DDPG+HER, Q-WSL attains an improvement of up to 5.48 percentage points. The average improvement across the five tasks is 37.645 percentage points. In summary, Q-WSL constitutes a straightforward yet highly effective approach that attains state-of-the-art performance in goal-conditioned reinforcement learning.
\subsection{Sample Efficiency}
\begin{wrapfigure}[17]{r}{0.5\linewidth}
   \centering
   \includegraphics[width=0.9\linewidth]{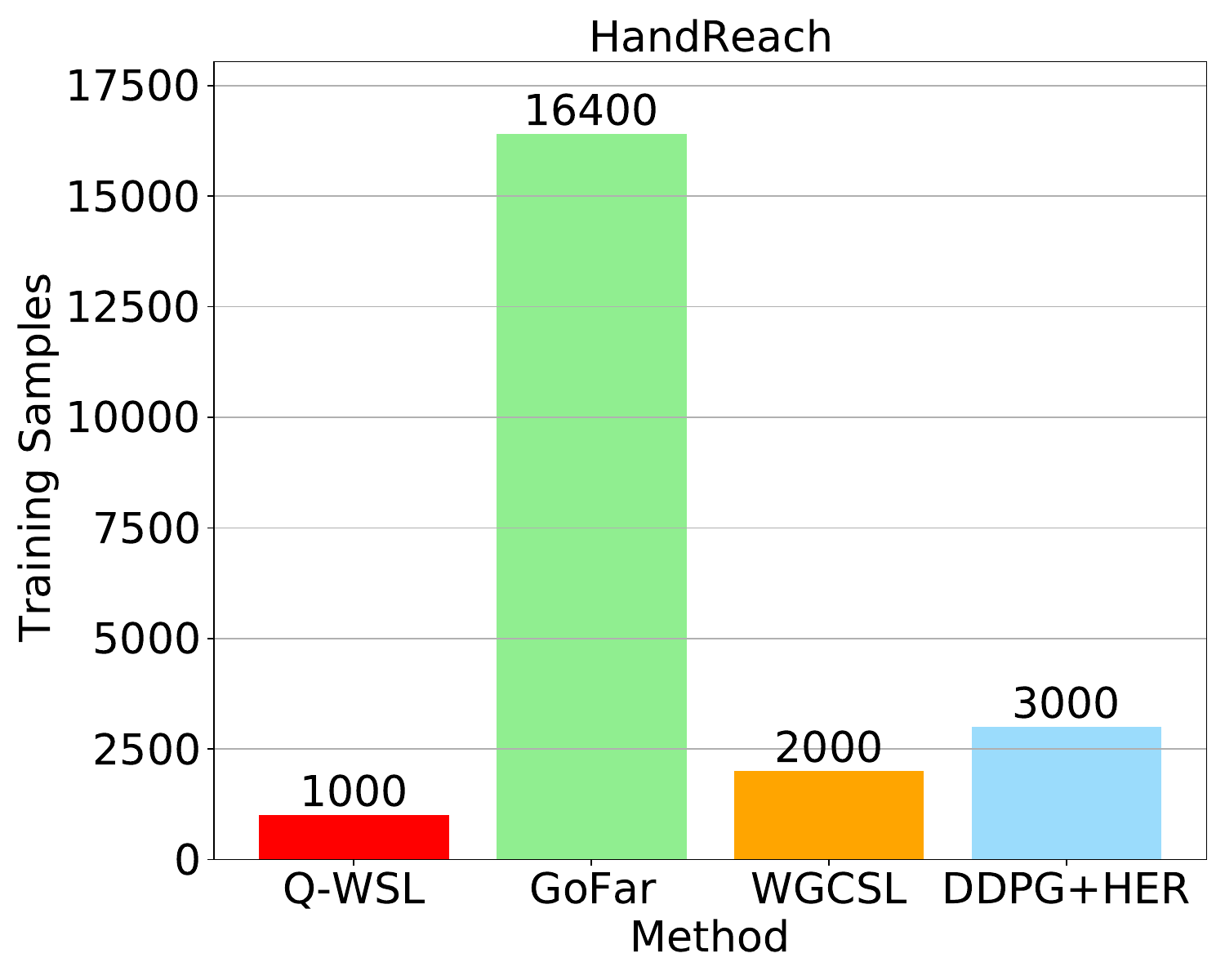} 
   \caption{\footnotesize
   Training samples for HandReach environment under 0.1 success rate.
   }
   \label{fig:multi_goal training samples results}
\end{wrapfigure}
To evaluate the sample efficiency of Q-WSL compared to baseline methods, we analyze the number of training samples (i.e., $\left\langle {s,a,g^{\prime},g} \right\rangle$ tuples, $g^{\prime}$ is the relabeled goal) required to attains 
a specific mean success rate. For this comparison, we have selected competitive baselines that employ goal relabeling strategies in the HandReach environment.
The results are depicted in \Cref{fig:multi_goal training samples results}. In the HandReach environment, as shown in \Cref{fig:multi_goal training samples results}, Q-WSL requires only 1000 samples to attains the 0.1 mean success rate, demonstrating superior sample efficiency.
In this case,
Q-WSL is at least twice the sample efficiency of previous works.
The reason why a state-of-the-art baseline DWSL is not listed here is that there is basically no success rate in the HandReach environment.
Overall, Q-WSL enhances sample efficiency by an average factor of seven (6.8) compared to the baseline methods.
\subsection{Robust to Environmental Stochasticity}
To test whether our Q-WSL is robust to random environmental factors, we follow Gofar's settings.
Specifically,
we examine a modified FetchPush environment characterized by the introduction of Gaussian noise with a zero mean before action execution. This modification generates various environmental conditions with standard deviations of ${0.2, 0.5, 1.0, 1.5}$, allowing us to analyze the robustness and performance of the proposed method under differing levels of stochasticity.

\begin{wrapfigure}[16]{r}{0.6\linewidth}
    \vspace{-0.5em}
    \centering
   \includegraphics[width=0.99\linewidth]{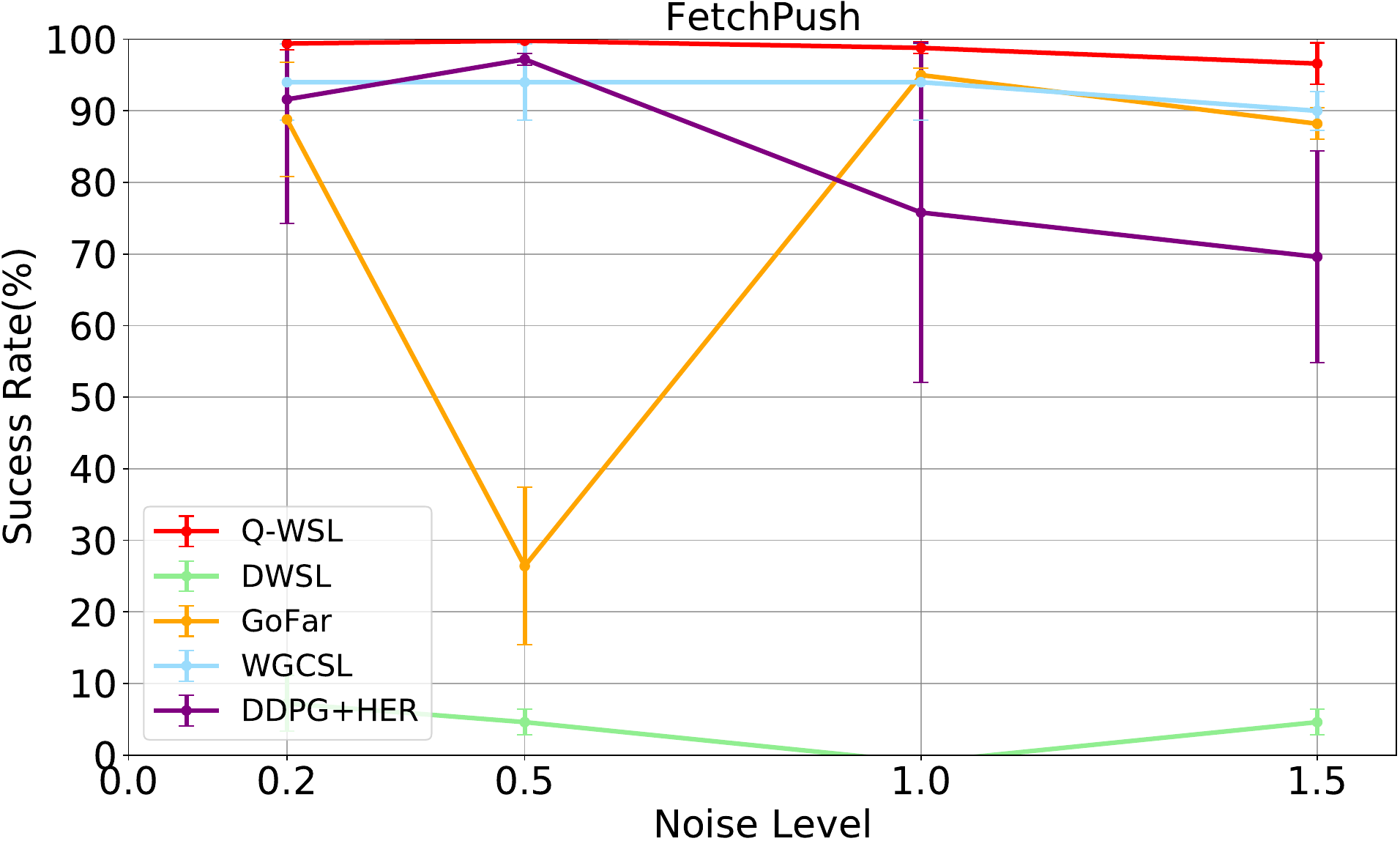}   
   \caption{\footnotesize
   Mean success rate (\%) for FetchPush task under environment stochasticity
   }
   \label{fig:multi_goal environment stochasticity results}
\end{wrapfigure}
As we see in \Cref{fig:multi_goal environment stochasticity results}, Q-WSL
is the most robust to stochasticity in the FetchPush environment, also outperforming 
baseline algorithms in terms of mean success rate under various noise levels.
WGCSL exhibits minimal sensitivity to variations in noise levels, whereas DDPG+HER
is more sensitive, and GoFar is the most affected by noise fluctuations. Despite DWSL's insensitivity to noise, its overall performance remains suboptimal.
We analyze that the assumption of deterministic dynamics in the self-supervised learning methods such as WGCSL, GoFar, and DWSL may lead to overly optimistic performance estimates in stochastic environments. In contrast, reinforcement learning methods are capable of effectively adapting to these changes.
\subsection{Robust to Reward Function}
Because the benchmark utilizes a default reward value of 
\Cref{eq:1}, we set it to $\mathbb{1}\left[\phi(s_i) = g\right]$ to assess the algorithm's robustness to variations in reward structure.
The comparison is shown in
\Cref{success 0_1 table}.
From
\Cref{success 0_1 table},
in the all 5 goal-conditioned RL benchmarks,
we can see Q-WSL still attains significantly higher performance than prior baselines and
HER principle is also still important to learning sparse reward. 
Moreover,
DDPG+HER is poor performance compared to GCWSL as it is always zero in indicator reward,
this means Q-learning is less robust to reward function.
This phenomenon and shortcomings of Q-learning have also been demonstrated in \citep{eysenbach2020c}.
\begin{table}[ht]
\renewcommand{\arraystretch}{1.5}
\centering
\caption{Mean success rate (\%) for challenging multi-goal robotics tasks under indicator reward after training}
\label{success 0_1 table}
\resizebox{\linewidth}{!}{
\begin{tabular}{l|llllll}
\hline
\textbf{Task}   & \textbf{Q-WSL(Ours)} & DWSL &GoFar & WGCSL & DDPG+HER & DDPG \\ \hline
FetchReach         & \colorbox{magiccolor}{100.0} $\pm$ 0.0              &98.0 $\pm$ 1.58                &100.0 $\pm$ 0.0                &100.0 $\pm$ 0.0              &100.0 $\pm$ 0.0              &100.0 $\pm$ 0.0                  \\
FetchPush         &\colorbox{magiccolor}{99.4} $\pm$ 0.89               &4.4 $\pm$ 1.8                & 9.4 $\pm$ 5.1               &83.0 $\pm$ 8.6              &57.4 $\pm$ 12.7                &4.4 $\pm$ 1.8              \\
FetchSlide         &\colorbox{magiccolor}{44.6} $\pm$ 6.3              &0.4 $\pm$ 0.5                &2.0 $\pm$ 2.1                &31.4 $\pm$ 4.3               &7.4 $\pm$ 1.9              &0.2 $\pm$ 0.4             \\
FetchPickAndPlace          &\colorbox{magiccolor}{45.4} $\pm$ 14.5               & 4.4 $\pm$ 1.5               &16.5 $\pm$ 9.2                &40.6 $\pm$ 6.1              &23.6 $\pm$ 2.6              &4.4 $\pm$ 1.7         \\
HandReach                         & \colorbox{magiccolor}{75.6} $\pm$ 5.68              &0.0 $\pm$ 0.0                &21.2 $\pm$ 4.5                &18.8 $\pm$ 10.4              &40.2 $\pm$ 12.7               &0.0 $\pm$ 0.0                 \\ \hline
\textbf{Average}                      & \colorbox{magiccolor}{73.0}               &21.44                &29.82                &54.76              & 45.72              & 21.8   \\ \hline
\end{tabular}
}
\end{table}
\subsection{Ablation Studies} \label{sec:6.2}
\begin{wrapfigure}[12]{r}{0.65\linewidth}
    \vspace{-0.5em}
    \begin{minipage}{\linewidth}
		\vspace{3pt}
		\centerline{\includegraphics[width=0.95\textwidth]{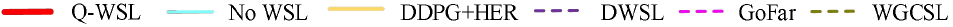}}
      
    \end{minipage}
    \begin{minipage}{0.495\linewidth}
		\vspace{3pt}
		\centerline{\includegraphics[width=\textwidth]{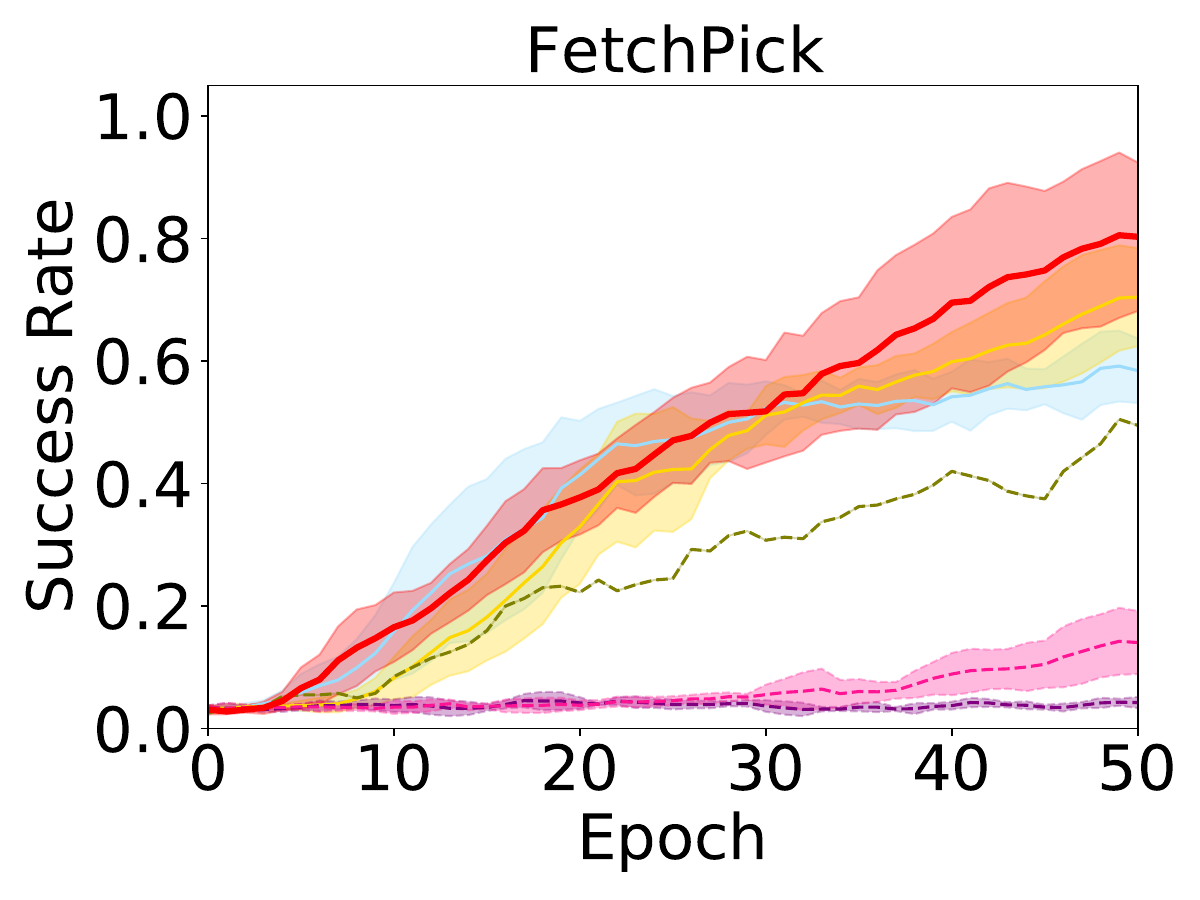}}
	\end{minipage}
    \begin{minipage}{0.495\linewidth}
		\vspace{3pt}
		\centerline{\includegraphics[width=\textwidth]{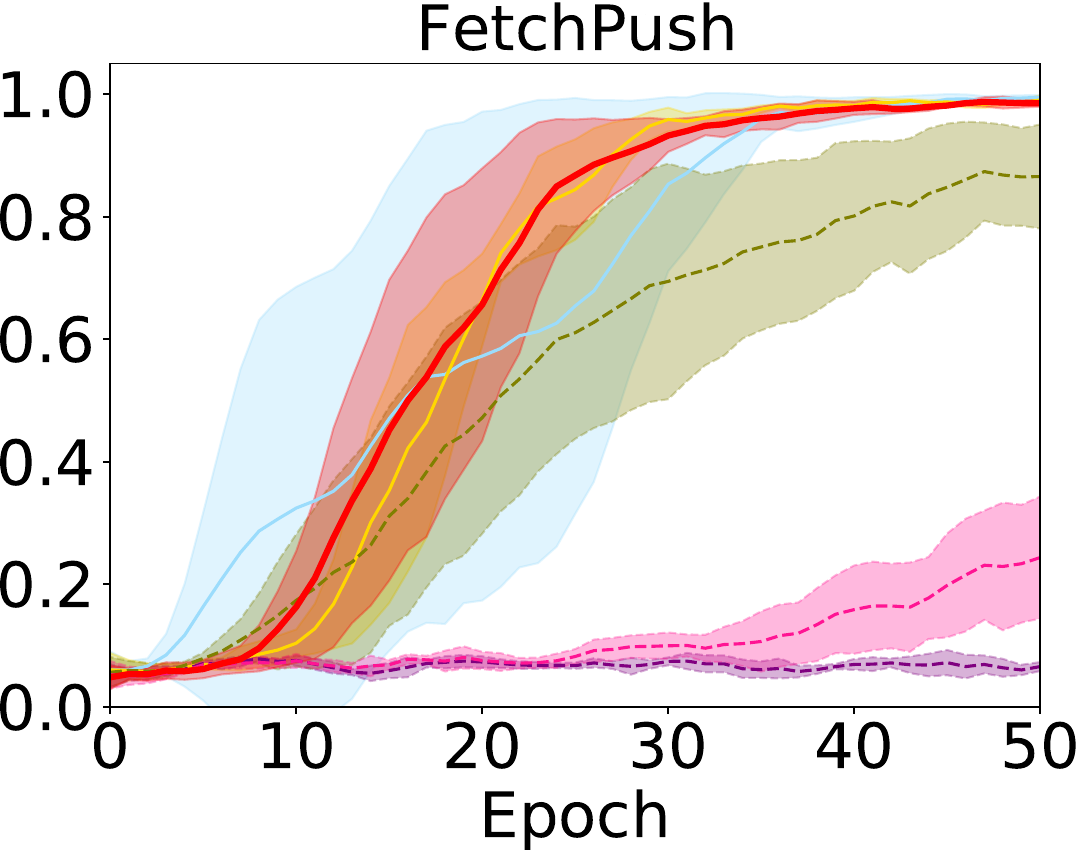}}
	\end{minipage}
    \caption{
    Ablation studies in FetchPick and FetchPush tasks.
    }
	\label{fig:multi_goal ablation results}
\end{wrapfigure}

To evaluate the significance of incorporating both Q-learning and advantage-weighted regression in Q-WSL, we conducted ablation experiments comparing various Q-WSL variants with HER. For the joint optimization in \Cref{eq:20}, the parameter $\eta$ is set to 0.1 by default.
We conduct our experiments under the following conditions:
\begin{itemize}
	\item \textbf{Q-WSL}
    Q-learning + Goal-conditioned Weighted Supervised Learning (GCWSL).
	\item \textbf{No Q-learning}
    which is equivalent to WGCSL.
	\item \textbf{No WSL}
    which is equivalent to Q-learning as described in \Cref{eq:18}.
\end{itemize}
Empirical results are shown in \Cref{fig:multi_goal ablation results}. 
As depicted in \Cref{fig:multi_goal ablation results}, Q-learning and GCWSL  are essential components. Q-WSL 
demonstrates faster learning than DDPG+HER, whereas the state-of-the-art baseline DWSL 

fails to learn in FetchPick and FetchPush, suggesting that supervised learning is sub-
optimal for relabeled data. The integration of Q-learning and GCWSL significantly enhances performance.

The underlying rationale for these results is that Q-learning and GCWSL mutually reinforce each other within the Q-WSL framework. GCWSL supplies high-quality samples for policy training, while the Q-learning refines the policy further, thereby guiding the curriculum of advantage-weighted regression.

\section{Conclusion}
In this paper, we develop a novel approach termed Q-learning Weighted Supervised Learning (Q-WSL) for Goal-Conditioned Reinforcement Learning (GCRL), which employs the principles of Dynamic Programming from Q-learning to overcome the notable shortcomings of the Goal-Conditioned Weighted Supervised Learning (GCWSL) technique. This innovative framework significantly enhances the capabilities of goal-conditioned reinforcement learning algorithms. Q-WSL specifically targets situations where GCWSL demonstrates suboptimal results due to the necessity of stitching trajectories for task completion. By utilizing Q-learning, our method effectively propagates optimal values backward to the initial states, thereby ensuring optimal decision-making.
Experimental evaluations in rigorous goal-conditioned RL scenarios show that Q-WSL not only outperforms existing advanced algorithms but also maintains consistent performance amidst variations in reward dynamics and environmental uncertainties, illustrating its robustness and versatility.

Future research will explore the potential of purely supervised learning and other approaches, such as efficient data augmentation or other Dynamic Programming former, to facilitate faster and more efficient learning of enhanced goal-conditioned policies.


\bibliographystyle{plainnat}
\bibliography{conferences, main}


\newpage

\appendix
\section{Appendix}

\subsection{PROOF of \Cref{theorem:3.2}} \label{sc:a.1}
Goal-conditioned AWR maximizes the expected improvement over a sampling policy and can implemented as a constrained policy search problem.
This optimization problem in goal-conditioned settings can be formulated as 
\citep{peng2020advantage,nair2020awac}:
\begin{align} \label{eq:6}
\arg\max_\pi=\int_g\rho_{\mathcal{B}_r}(g)\int_sd_{\mathcal{B}_r}(s|g)\int_a\left[A^\pi(s, a, g)\right]\boldsymbol{d}a\boldsymbol{d}s\boldsymbol{d}g,&\int_a\pi(a|s,g)\boldsymbol{d}a=1,\nonumber\\
&\mathbf{s.t.}~\operatorname{KL}(\pi_\theta\|\pi_{relabel})\leq\epsilon,
\end{align}
where $\rho_{{\mathcal{B}_r}}(g)$ represents the goal distribution in relabeled data, and 
$d_{\mathcal{B}_r}(s|g)=
\sum_{t=0}^{\inf}\gamma^tp(s_t=s|\pi_{relabel},g)$ denotes the goal-conditioned un-normalized discounted state distribution induced by the relabeling policy 
$\pi_{relabel}$ 
\citep{sutton2018reinforcement}. In this context, 
$\pi_{\psi}$ is the learned policy, ${\mathcal{B}_r}$ refers to the relabeled data, and $\pi_{relabel}$ signifies the behavioral policy responsible for collecting the dataset. KL represents the KL-divergence measurement metric, and $\epsilon$ is the threshold. By enforcing the Karush-Kuhn-Tucker (KKT) condition on \Cref{eq:6}, we obtain the Lagrangian:
\begin{align}
\mathcal{L}(\lambda,\pi,\alpha)&=\int_{g}\rho_{{\mathcal{B}_r}}(g)\int_{s}d_{{\mathcal{B}_r}}(s|g)\int_{a}\pi(a|s,g)\left[A^{\pi}(s,a,g)\right]
\boldsymbol{d}a\boldsymbol{d}s\boldsymbol{d}g \nonumber\\
&+\lambda(\epsilon-\int_g\rho_{\mathcal{B}_r}(g)\int_sd_{\mathcal{B}_r}(s|g)\operatorname{KL}(\pi_\psi\|\pi_{relabel})\boldsymbol{d}s)\boldsymbol{d}g \nonumber\\
&+\int_s\alpha(1-\int_a\pi(a|s,g)\boldsymbol{d}a)\boldsymbol{d}s.
\end{align}
Differentiating with respect to $\pi$:
\begin{align}
\frac{\partial\mathcal{L}}{\partial\pi}&=p_{\mathcal{B}_r}(s,g)A^\pi(s,a,g)-\lambda\cdot p_{\mathcal{B}_r}(s,g)\cdot\log\pi_{relabel}(a|s,g)\nonumber\\
&+\lambda p_{\mathcal{B}_r}(s,g)\log\pi(a|s,g)+\lambda p_{\mathcal{B}_r}(s,g)-\alpha,
\end{align}
where $p_{\mathcal{B}_r}(s,g)=\rho_{\mathcal{B}_r}(g)d_{\mathcal{B}_r}(s|g)$.
Then set $\frac{\partial\mathcal{E}}{\partial\pi}$ to zero we can have:
\begin{equation}
\pi(a|s,g)=\pi_{relabel}(a|s,g)\exp(\frac{1}{\lambda}A^\pi)\exp(-\frac{1}{p_{\mathcal{B}_r}(s,g)}\frac{\alpha}{\lambda}-1),
\end{equation}
where $\exp(-\frac1{p_{\mathcal{B}_r}(s,g)}\frac\alpha\lambda-1)$ is the partition function $Z(s, g)$ which normalizes the action distribution under state-goal
\citep{peng2020advantage}:
\begin{equation}
Z(s,g)=\exp(-\frac{1}{p_{\mathcal{B}_r}(s,g)}\frac{\alpha}{\lambda}-1)\nonumber\\
=\int_a^{\prime}\pi_{\mathcal{B}_r}(a'|s,g)\exp(\frac{1}{\lambda}A^{\pi}(s,a',g))\boldsymbol{d}a'.
\end{equation}
The closed-form solution can thus be expressed as follows:
\begin{equation}
\pi^*(a|s,g)=\frac{1}{Z(s,g)}\pi_{relabel}(a|s,g)\exp(\frac{1}{\lambda}A^\pi(s,a,g))
\end{equation}
Finally, since $\pi$ is a parameterized function, translate the solution into these policies, resulting in the following optimization objective:
\begin{align}
\underset{\pi}{\mathrm{arg}\operatorname*{min}}\mathbb{E}_{s,g\sim {\mathcal{B}_r}}\left[\mathrm{KL}(\pi^{*}(\cdot|s,g)\|\pi(\cdot|s,g))\right]
& =\arg\underset{\pi}{\operatorname*{min}}\mathbb{E}_{s,g\sim {\mathcal{B}_r}}\big[\mathrm{KL}(\frac{1}{Z(s,g)}\pi_{relabel}(a|s,g)\nonumber\\
&\cdot\exp(\frac{1}{\lambda}A^{\pi}(s,a,g))\|\pi(\cdot|s,g))\big]  \nonumber\\
&=\arg\min_{\pi}\mathbb{E}_{s,g\sim {\mathcal{B}_r}}\mathbb{E}_{a\sim {\mathcal{B}_r}}\big[-\pi(a|s,g)\nonumber\\
&\cdot\exp(\frac{1}{\lambda}A^{\pi}(s,a,g))\big]  
\end{align}
The objective can now be interpreted as a weighted maximum likelihood estimation. If the policy is deterministic, the objective can be simplified to:
\begin{equation} \label{eq:13}
\underset{\pi}{\arg\min}\mathbb{E}_{s,a,g\sim {\mathcal{B}_r}}\left[\exp(\frac{1}{\lambda}A^{\pi}(s,a,g)\cdot\|\pi_{\theta}(s,g)-a\|_2^2\right].
\end{equation}
It can be rewritten to maximize the objective form:
\begin{equation}
\mathcal{J}_{wbc}(\pi)=\mathbb{E}_{s,a,g\sim {\mathcal{B}_r}}\left[\exp(\frac{1}{\lambda}A^{\pi}(s,a,g)\cdot\log(\pi(a|s,g)))\right].
\end{equation}
However, $Z(s)$ is typically ignored in practical implementations \citep{peng2020advantage, nair2020awac}. Notably, \citep{nair2020awac} suggests that including $Z(s, g)$ often degrades performance. Therefore, for practical purposes, we also omit the term $Z(s)$.
If we rewrite $\exp(\frac{1}{\lambda}A^{\pi}(s,a,g)$ to a general function $f(A)$,
then GCWSL is the variant form of goal-conditioned AWR which guarantees policy improvement.

\subsection{PROOF of \Cref{theorem:5.1}} \label{sc:a.2}
Before proving, we note the following premises:
$1\geq \gamma \textgreater 1$, $1\geq\eta\textgreater0$, $\exp_{clip}(
A(s_{t},a_{t},g)\cdot\epsilon(A(s_{t},a_{t},g))\geq 1,g=\phi(s_i),i\geq t.$
\begin{align}
\mathcal{J}_{Q-WSL}&=\mathcal{J}_{q}+\eta\mathcal{J}_{wgcsl}\nonumber\\
&=\mathbb{E}_{g\sim p(g),\tau\sim\pi_{b}(\cdot|g),t\sim[1,T],i\sim[t,T]}\big[\gamma^{i-t}r(s_{t},a_{t},g)+\eta\gamma^{i-t}\nonumber\\
&\exp_{clip}(A(s_{t},a_{t},g))\cdot\epsilon(A(s_{t},a_{t},g))\log\pi(a_t\mid s_t,g)\big]\nonumber\\ 
&\geq\mathbb{E}_{g\sim p(g),\tau\sim\pi_{b}(\cdot|g),t\sim[1,T],i\sim[t,T]}\big[\exp_{clip}(
A(s_{t},a_{t},g))\nonumber\\
&\cdot\epsilon(A(s_{t},a_{t},g))\log\pi(a_t\mid s_t,g)\big]\nonumber\\
&\geq\mathbb{E}_{g\sim p(g),\tau\sim\pi_{b}(\cdot|g),t\sim[1,T],i\sim[t,T]}\big[\exp_{clip}(
A(s_{t},a_{t},g))\nonumber\\
&\cdot\epsilon(A(s_{t},a_{t},g))\log\pi(a_t\mid s_t,g)\big]\triangleq J_{WGCSL}(\pi)\nonumber\\
&\geq\mathbb{E}_{g\sim p(g),\tau\sim\pi_{b}(\cdot|g),t\sim[1,T],i\sim[t,T]}\big[\log\pi(a_t\mid s_t,g)\big]\nonumber\\
&\triangleq J_{GCSL}(\pi).
\end{align}

\subsection{Baseline Details} \label{sc:a.2}
In our experiments, all algorithms share the same actor-critic architecture 
(i.e, DDPG)
and hyperparameters. 
Throughout the data collection phase governed by policy $\pi$, Gaussian noise is systematically added with a mean of zero and a constant standard deviation, as detailed in \citep{fujimoto2018addressing}, to enhance exploration capabilities.
Implementations of DDPG, AM, MHER, DDPG+HER, and GCSL were sourced from GoFar's open code. Although WGCSL, GoFar, and DWSL are designed as offline algorithms, they are also applicable in off-policy settings; hence, we re-implemented them into our framework based on the principles outlined in the literature. Here, we specifically focus on methods based on advantage-weighted regression.
Before introducing the methods, we will first define $\mathcal{B}$ as the replay buffer and $\mathcal{B}_r$ represent the data relabeled from this buffer. We will evaluate the following GCWSL methods for comparison.
\begin{itemize}
    \item \textbf{GCSL}
    \citep{ghosh2021learning}
    is implemented by removing the critic component of the DDPG algorithm and modifying the policy loss to be based on maximum likelihood estimation:
    \begin{equation}
    \min_{\pi}-\mathbb{E}_{(s,a,g)\sim \mathcal{B}_r}\left[\log\pi(a\mid s,g)\right].  
    \end{equation}
    GCSL can be regarded as GCWSL when $f(A)=1$. 
    The following baselines are all on the offline goal-conditioned.
    We re-implement them on the online.
    \item \textbf{WGCSL}
    \citep{yang2022rethinking}
    is implemented as an extension of GCSL by incorporating a Q-function.
    The training of the Q-function employs Temporal Difference (TD) error, similar to the methodology used in the Deep Deterministic Policy Gradient (DDPG) algorithm. Furthermore, this process is enhanced by incorporating advantage weighting into the regression loss function.
    The advantage term that we compute is denoted as $A(s_t,a_t,g)=r(s_t,a_t,g)+\gamma Q(s_{t+1},\pi(s_{t+1},g),g)-Q(s_t,a_t,g)$.
    By employing this denotes,
    the policy objective of WGCSL is:
    \begin{equation}
    \min_{\pi}-\mathbb{E}_{(s_t,a_t,g)\thicksim \mathcal{B}_r}\big[\gamma^{i-t}\exp_{clip}(A(s_{t},a_{t},g))\nonumber\\
    \cdot\epsilon(A(s_{t},a_{t},g))
    \log\pi(a_t\mid s_t,g)\big],  
    \end{equation}
    where $\exp_{clip}$ is the clip for numerical stability
    and 
    \begin{align}
    \epsilon(A(s_t,a_t,g))=\begin{cases}1,&A(s_t,a_t,g)>\hat{A} \\\epsilon_{min},&\text{otherwise}\end{cases}, &
    \end{align} ($\hat{A}$ is a threshold, $\epsilon_{min}$ is a small positive value).
    \item \textbf{GoFar}
    \citep{ma2022offline}
    proposed the adoption of a state-matching objective for multi-goal RL,
    wherein a reward discriminator and a bootstrapped value function were employed to assign weights to the imitation learning loss.
    The policy objective of GoFar is:
    \begin{equation}
    \min_{\pi}-\mathbb{E}_{(s_t,a_t,g)\sim \mathcal{B}}[\max(A(s,a,g)+1,0)\log\pi(a_t|s_t,g)].
    \end{equation}
    The advantage function is estimated using discriminator-based rewards. The discriminator, denoted as $c$, is trained to minimize
    \begin{equation}
    \mathbb{E}_{g\sim p(g)}[\mathbb{E}_{p(s;g)}\big[\log c(s,g)\big]+\mathbb{E}_{(s,g)\sim \mathcal{B}}[\log(1-c(s,g))]\big]
    \end{equation}.
    The value function $V$ is trained to minimize
    $(1-\gamma)\mathbb{E}_{(s,g)\thicksim\mu_0,p(g)}[V(s,g)]+\frac12\mathbb{E}_{(s,a,g,s^{\prime})\thicksim \mathcal{B}}[(r(s;g)+$\\
    $\gamma V(s^{\prime};g)-V(s;g)+1)^2]$,
    where $V\geq0$. After obtaining the optimal $V^{*}$, the calculation method of advantage function $A(s,a,g)$ is the same as WGCSL. We have re-implemented GoFar with an enhanced version, GoFar+HER
    (i.e, $g= g$), to attain higher sample efficiency.
    \item \textbf{DWSL}
    \citep{hejna2023distance} presents a cutting-edge supervised learning approach that models the empirical distribution $p_{\zeta}^r$ of discrete state-to-goal distances $\hat{d}$
    based on offline data. We adapted this method to an off-policy setting, implementing the following principle for distribution training:
    \begin{equation}
    \max_\zeta\mathbb{E}_{\mathcal{B}_r}\left[\log p_\zeta^r\left((j-i-1)//N|s_i,g\right)\right],
    \end{equation}
    where $s_i,s_{i+1},g=\phi(s_j),j>i$ are the relabeled data from the $\mathcal{B}_r$, and $N$ represents the number of steps.
    DWSL calculates significant statistics from the distance distribution to determine the shortest path estimates between any two states, thus avoiding the challenges of bootstrapping. Specifically, DWSL utilizes the LogSumExp function for a smooth estimation of the minimum distance.The method calculates distances between any two states as follows:
    \begin{equation}
    \hat{d}(s_i,g)=-\alpha\log\mathbb{E}_k\sim p_\zeta^r(\cdot|(s_i,g)\left[e^{-k/(I\alpha)}\right]
    \end{equation}
    and
    \begin{equation}
    \hat{d}(s_{i+1},g)=-\alpha\log\mathbb{E}_k\sim p_\zeta^r(\cdot|(s_{i+1},g)
    \cdot\left[e^{-k/(I\alpha)}\right],
    \end{equation}
    where $I$ is the bin count in the distribution and 
    $\alpha$ is the temperature. We follow DWSL to set $I$ to default to 100.

    Ultimately, DWSL reweights actions according to their effectiveness in reducing the estimated distance to the goal:
    \begin{equation}
    A=\hat{d}(s_i,\phi(s_j))-c(s_i,g)-\hat{d}(s_{i+1},g),
    \end{equation}
    where $c(s_i,g)=\mathbb{1}\left[\phi(c_{i+1})\neq g\right]/I$.

   The policy is then extracted as:
   \begin{equation}
   \max_{\theta}\mathbb{E}_{\mathcal{B}_r}\left[e^{A/\omega}\log\pi_{\theta}(a_i|s_i,g)\right],
   \end{equation}
   where $\omega$ is the temperature parameter, and $\pi$ is the policy parameterized by $\theta$.
\end{itemize}
\subsection{Additional Experiments Results}
In this section, we examine the robustness of Q-WSL concerning various parameters, including the relabeling ratio. Due to space limitations, detailed ablation studies are provided here instead of the main text.

\begin{figure}[h]
    \begin{minipage}{\linewidth}
		\vspace{3pt}
		\centerline{\includegraphics[width=0.5\textwidth]{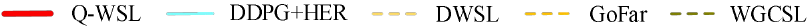}}
	\end{minipage}
 
    \begin{minipage}{0.245\linewidth}
		\vspace{3pt}
		\centerline{\includegraphics[width=\textwidth]{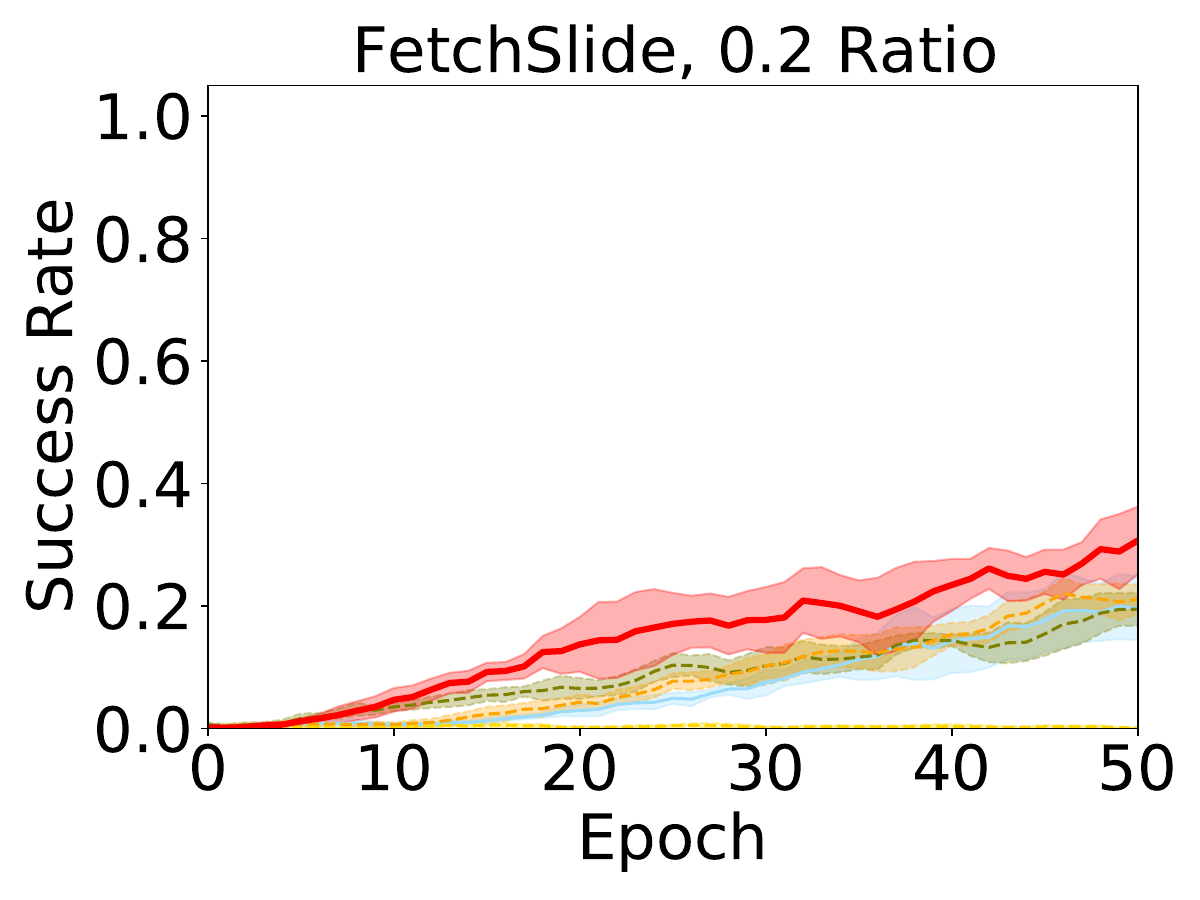}}
	\end{minipage}
    \begin{minipage}{0.245\linewidth}
		\vspace{3pt}
		\centerline{\includegraphics[width=\textwidth]{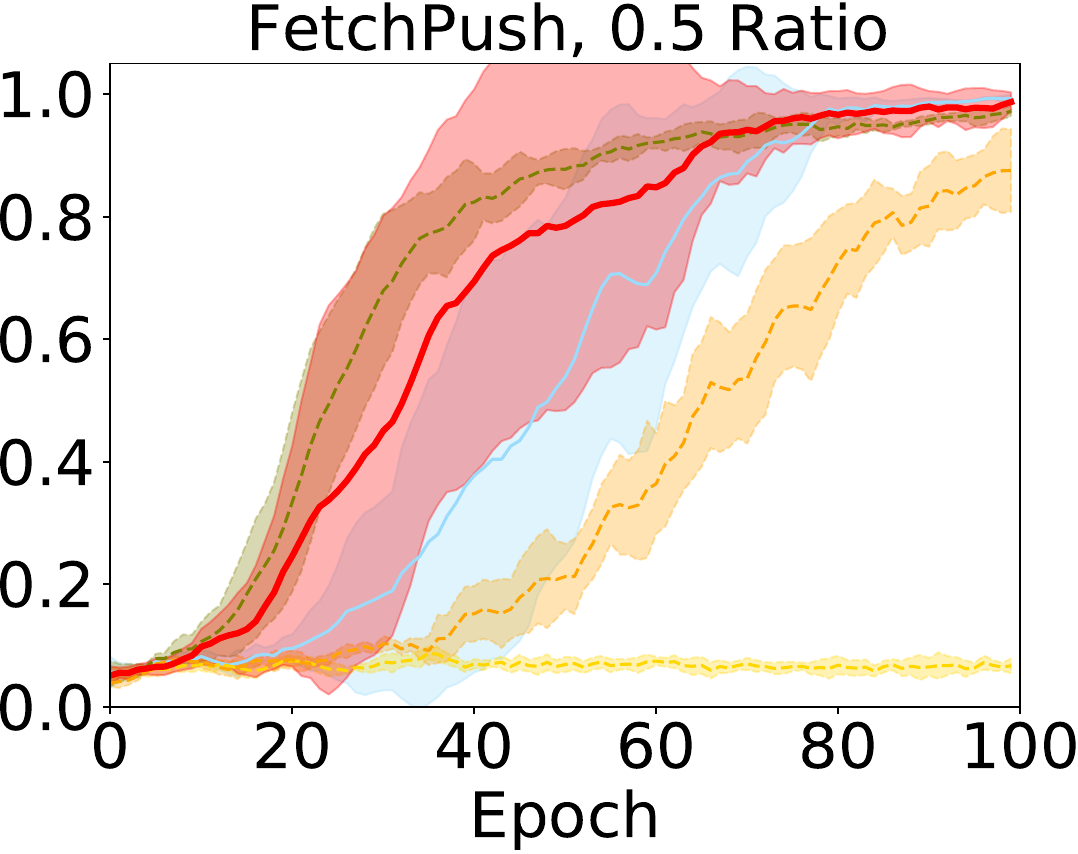}}
	\end{minipage}
	\begin{minipage}{0.245\linewidth}
		\vspace{3pt}
		\centerline{\includegraphics[width=\textwidth]{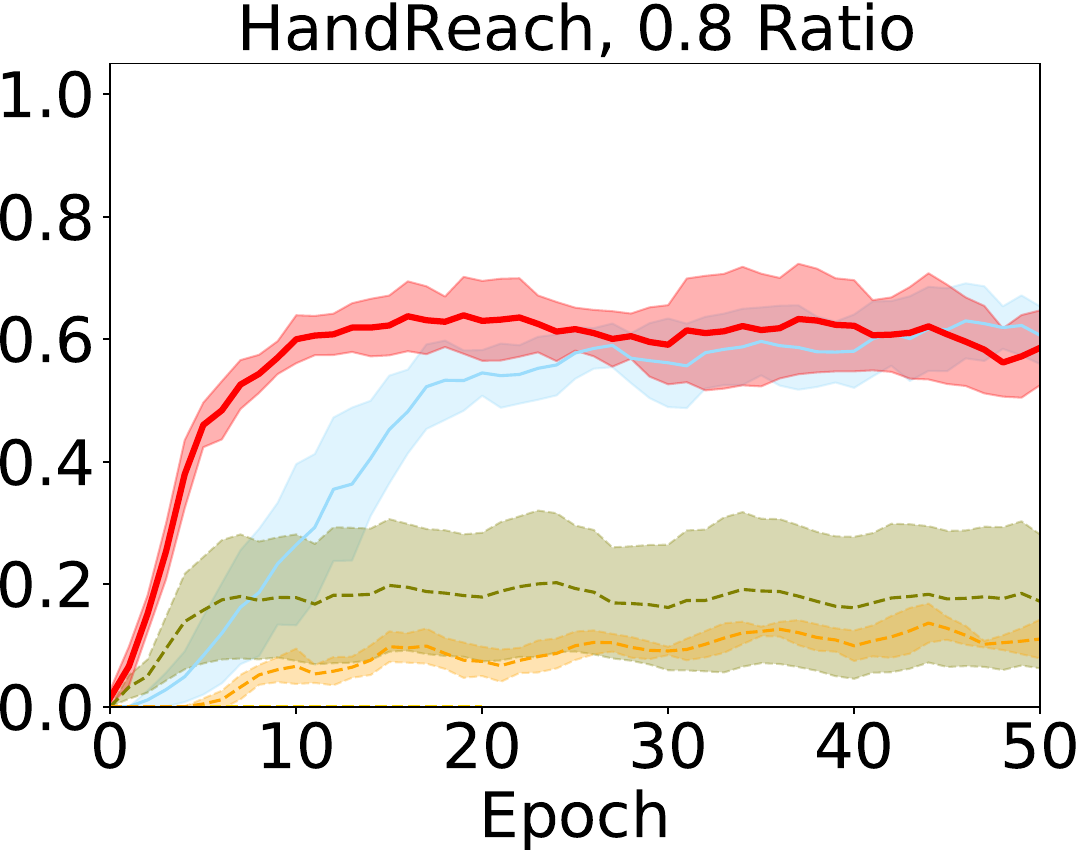}}
	\end{minipage}
    \begin{minipage}{0.245\linewidth}
		\vspace{3pt}
		\centerline{\includegraphics[width=\textwidth]{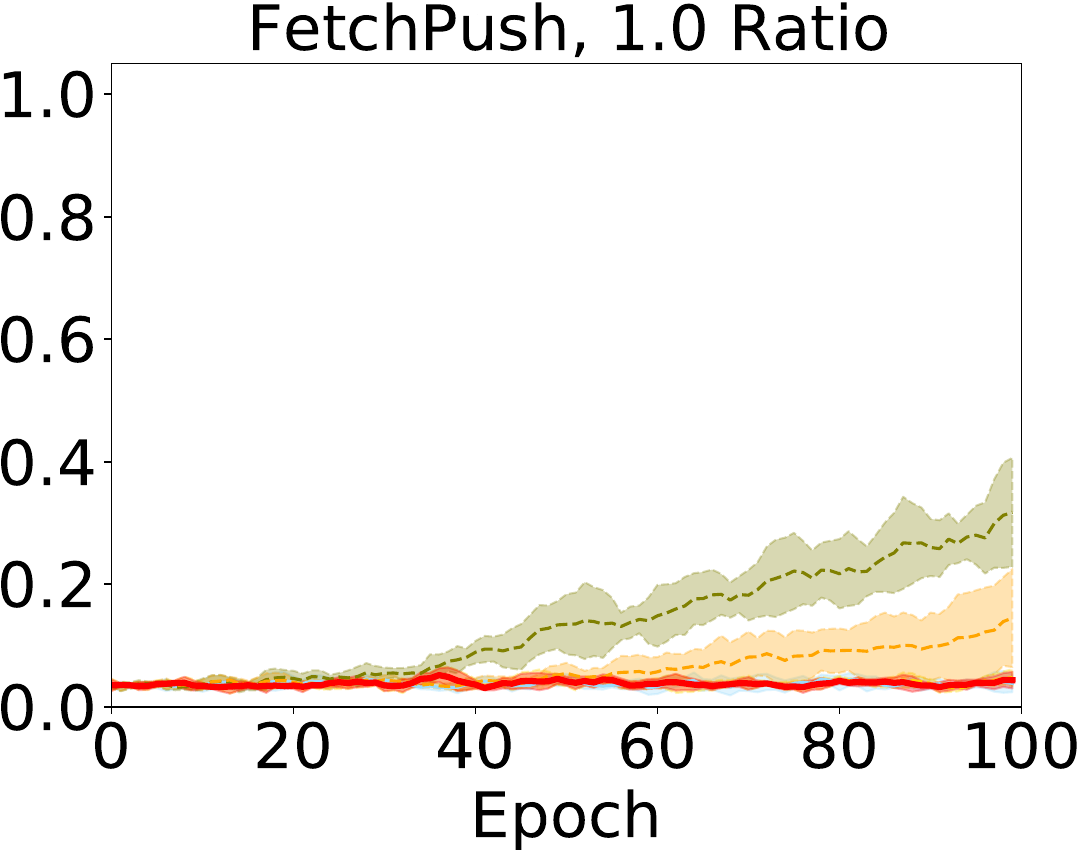}}
	\end{minipage}
    \caption{
    Relabel ratio ablation studies in such multi-goal tasks. The results demonstrate that Q-WSL outperforms other competitive HER-based methods, except in cases where the labeling rate is 1.0.
    }
	\label{fig:multi_goal relabel_rate results}
\end{figure}

\paragraph{Relabeling Ratio} The multi-goal framework we consider assumes that data is associated with goal labels. We explore the influence of the relabeling ratio on performance, as depicted in \Cref{fig:multi_goal relabel_rate results}. The results in \Cref{fig:multi_goal relabel_rate results} indicate that Q-WSL remains robust across different relabeling ratios. Furthermore, Q-WSL consistently outperforms competitive methods such as WGCSL and DDPG+HER at various relabeling ratios.

\begin{figure}[h]
    \begin{minipage}{\linewidth}
		\vspace{3pt}
		\centerline{\includegraphics[width=0.8\textwidth]{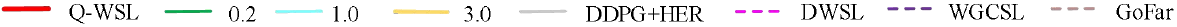}}
	\end{minipage}
 
    \begin{minipage}{0.245\linewidth}
		\vspace{3pt}
		\centerline{\includegraphics[width=\textwidth]{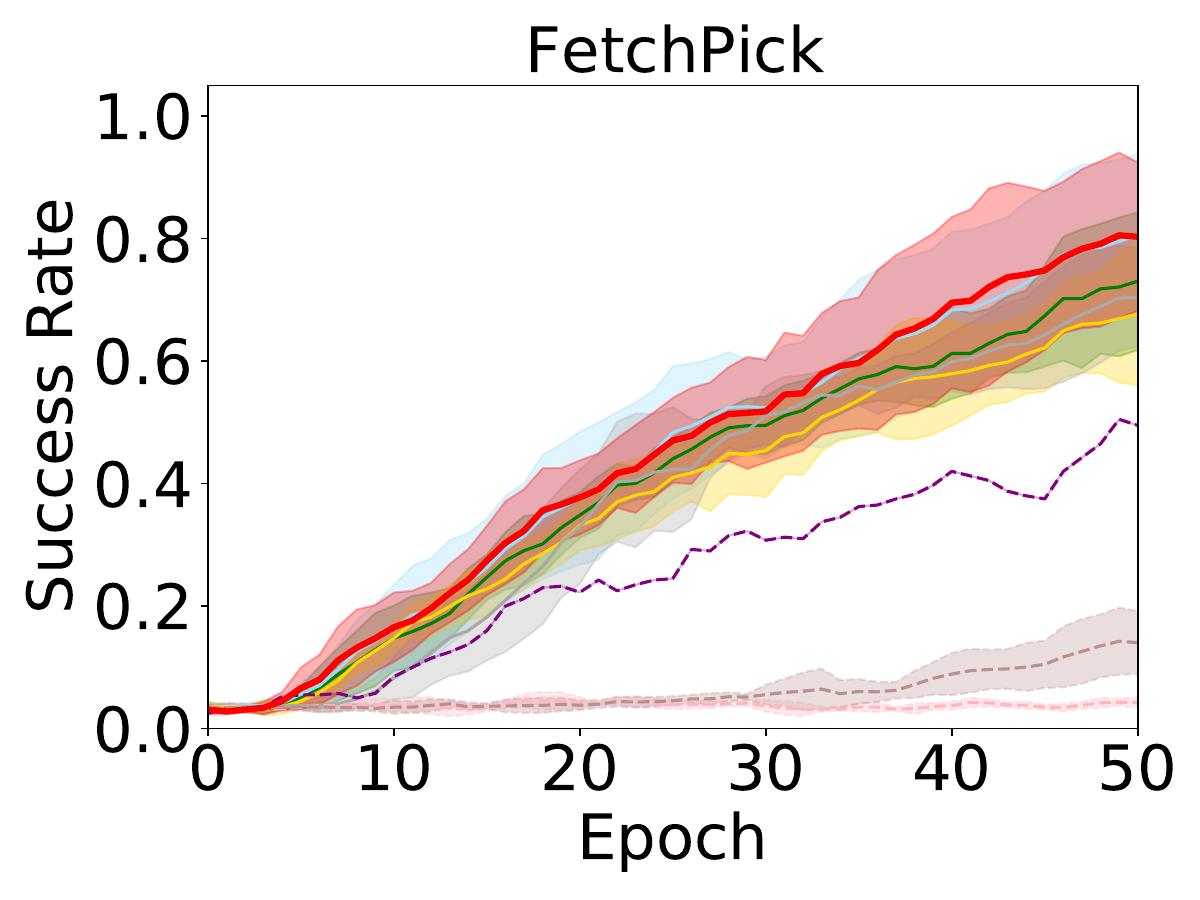}}
	\end{minipage}
    \begin{minipage}{0.245\linewidth}
		\vspace{3pt}
		\centerline{\includegraphics[width=\textwidth]{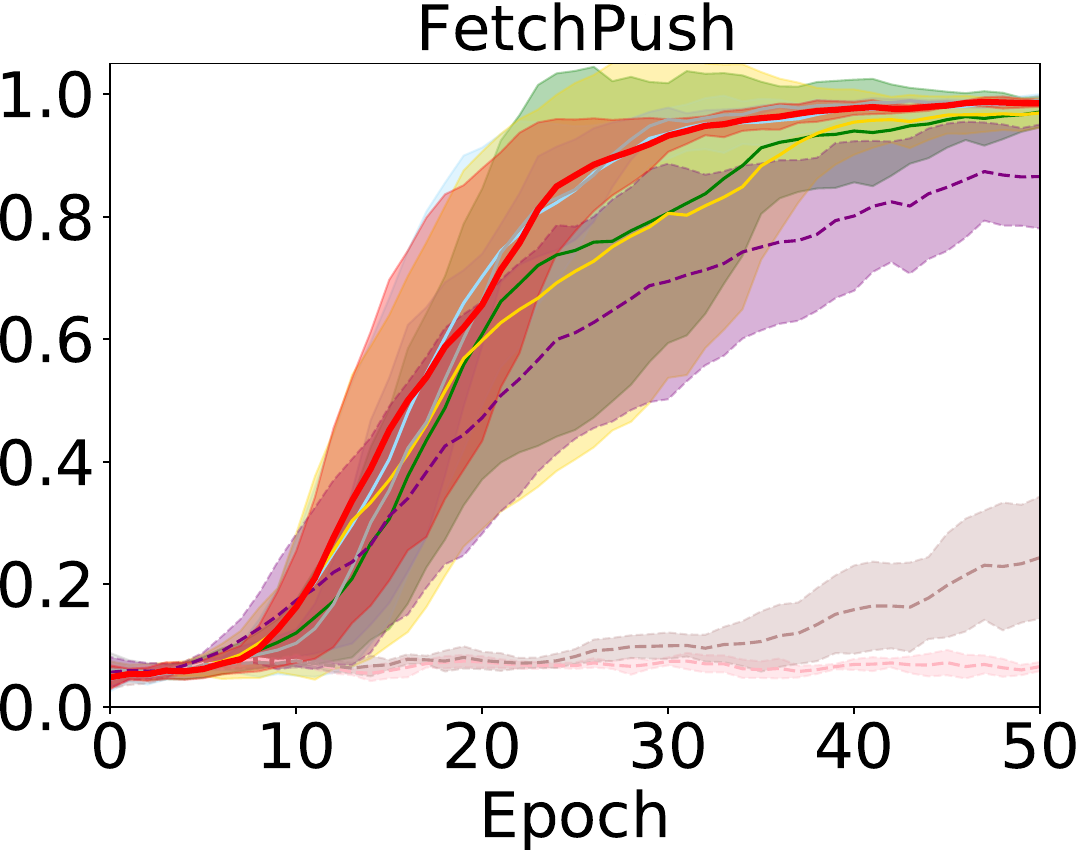}}
	\end{minipage}
	\begin{minipage}{0.245\linewidth}
		\vspace{3pt}
		\centerline{\includegraphics[width=\textwidth]{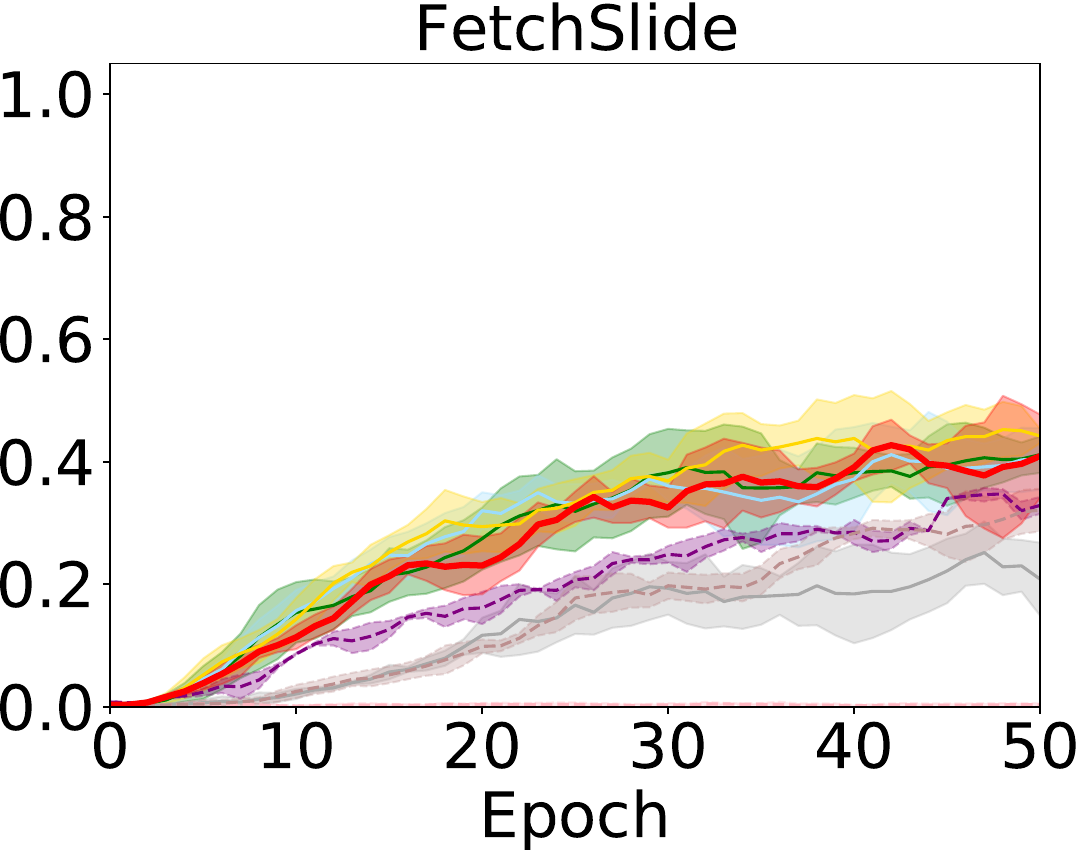}}
	\end{minipage}
    \begin{minipage}{0.245\linewidth}
		\vspace{3pt}
		\centerline{\includegraphics[width=\textwidth]{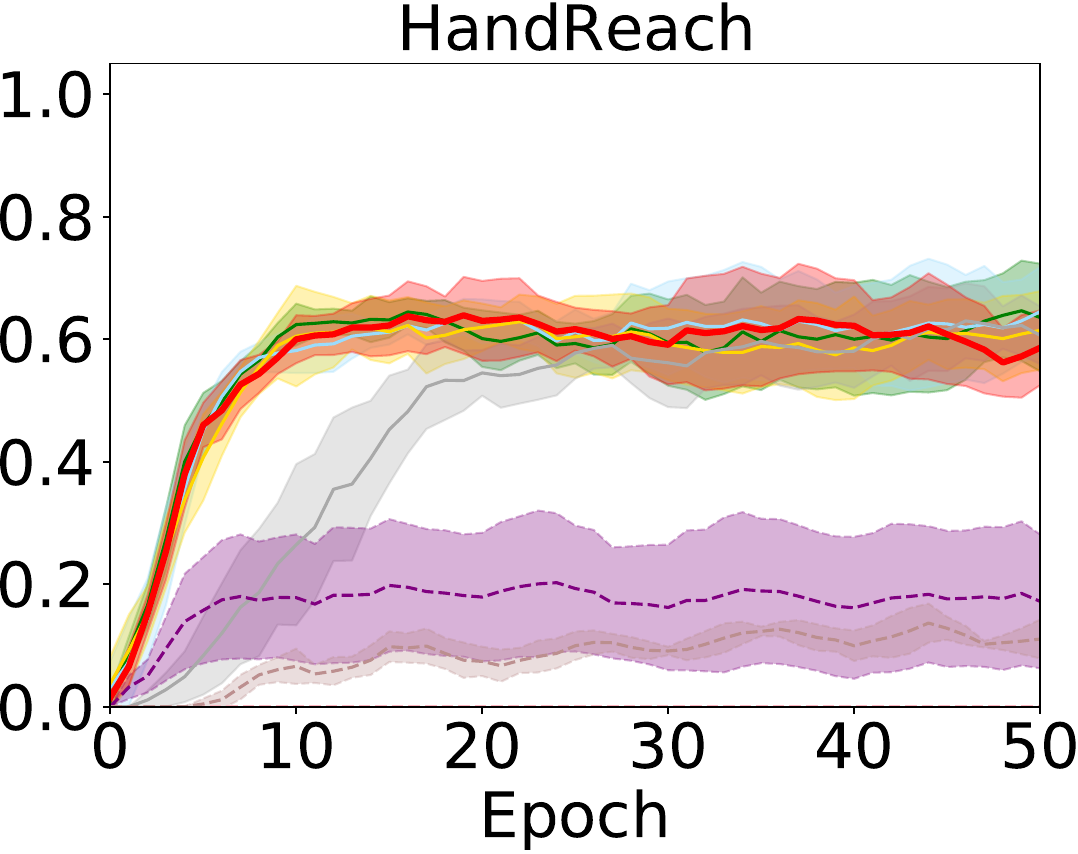}}
	\end{minipage}
    \caption{
    Hyperparameter $\eta$ ablation studies in such multi-goal tasks. Although the parameter $\eta$ can take different values, the results demonstrate that Q-WSL remains unaffected by variations in $\eta$, consistently achieving state-of-the-art performance.
    }
	\label{fig:multi_goal parameter results}
\end{figure}

\paragraph{The Impact of Hyperparameter $\eta$} Since our approach optimizes both supervised learning (SL) and reinforcement learning (RL) concurrently, this section explores the influence of the balancing parameter $\eta$. We evaluate $\eta$ values from the set $\{0.1,0.2,1.0,3.0\}$ and compare the results against competitive HER-based algorithms such as WGCSL and DDPG+HER, as shown in \Cref{fig:multi_goal parameter results}. The findings in \Cref{fig:multi_goal parameter results} reveal that Q-WSL consistently delivers superior performance over the other algorithms, regardless of the 
$\eta$ parameter variation. This demonstrates that our method maintains robustness and is not significantly affected by changes in the $\eta$ parameter.
\subsection{Hyperparameters} \label{appendix:a.7}
For all experiments, we utilize the Adam optimizer. To relabel goals, we uniformly sample from all future states within each trajectory. For baseline methods employing discount factors, we set 
$\gamma = 0.98$ across all Gym Robotics environments. Each algorithm is configured with a consistent set of hyperparameters for all multi-goal tasks. The hyperparameters for DWSL, GoFar, WGCSL, GCSL, MHER, and DDPG have been optimized for our task set, and we use the values reported in previous studies. In our implementation of AM, we use the same network architecture as DDPG and therefore adopt its hyperparameters.
\begin{table}[H]
    \centering
    \footnotesize 
    \caption{Hyperparameters for Baselines.}
    \label{table:mt-hyperparameters}
    \begin{tabular}{p{3.5cm}|p{3.0cm}} \toprule
        Actor and critic networks & value \\ \midrule 
        Learning rate & 1e-3 \\
        Initial buffer size & $10^6$ \\
        Polyak-averaging coefficient & 0.95\\
        Action L2 norm coefficient & 1.0 \\
        Observation clipping & [-200,200] \\
        Batch size & 256 \\
        Rollouts per MPI worker & 2 \\
        Number of MPI workers & 16\\
        Epochs & 50\\
        Cycles per epoch & 50\\
        Batches per cycle & 40\\
        Test rollouts per epoch & 10\\
        Probability of random actions & 0.3\\
        Gaussian noise & 0.2\\
        Probability of hindsight experience replay & 0.8\\
        Normalized clipping & [-5, 5]\\
        $\eta$ & 0.1\\
        \bottomrule
    \end{tabular}
\end{table}
All hyperparameters are described in greater detail in \citep{andrychowicz2017hindsight}.

\end{document}